\documentclass[10pt,twocolumn,letterpaper]{article}

\usepackage[pagenumbers]{iccv} 

\usepackage[accsupp]{axessibility}
\usepackage{comment}
\usepackage{soul}
\usepackage{graphicx} 
\usepackage{multirow} 
\usepackage{array} 

\definecolor{iccvblue}{rgb}{0.21,0.49,0.74}
\usepackage[pagebackref,breaklinks,colorlinks,allcolors=iccvblue]{hyperref}

\usepackage{amsmath}
\usepackage{pifont}
\usepackage[dvipsnames]{xcolor}
\newcommand{\cmark}{\textcolor{ForestGreen}{\ding{51}}}%
\newcommand{\xmark}{\textcolor{BrickRed}{\ding{55}}}%

\DeclareMathOperator*{\argmax}{arg\,max}

\newcommand\blankfootnote[1]{%
  \begingroup
  \renewcommand\thefootnote{}\footnote{#1}%
  \addtocounter{footnote}{-1}%
  \endgroup
}

\def\paperID{9957} 
\def\confName{ICCV}
\def\confYear{2025}

\title{Prior2Former - Evidential Modeling of Mask Transformers for\\Assumption-Free Open-World Panoptic Segmentation}

\newcommand{\theauthors}{
Sebastian Schmidt$^{1,2,*}$ \quad Julius Körner$^{1,2,*}$ \quad Dominik Fuchsgruber$^1$\\
Stefano Gasperini$^{1,3,5}$ \quad Federico Tombari$^{1,4}$ \quad Stephan Günnemann$^{1,5}$\\[0.5em]
$^1$ Technical University of Munich \quad 
$^2$ BMW Group \quad 
$^3$ Visualais \\
$^4$ Google \quad 
$^5$ Munich Center for Machine Learning

}

\author{\theauthors}

\begin{document}
\maketitle

\blankfootnote{$^*$ Equal Contribution}
\blankfootnote{Corresponding author: {\tt sebastian95.schmidt@tum.de} }
\blankfootnote{Project page: \text{\url{
www.cs.cit.tum.de/daml/prior2former}}}

\begin{abstract}
In panoptic segmentation, individual instances must be separated within semantic classes. 
As state-of-the-art methods rely on a pre-defined set of classes, they struggle with novel categories and out-of-distribution (OOD) data. This is particularly problematic in safety-critical applications, such as autonomous driving, where reliability in unseen scenarios is essential.
We address the gap between outstanding benchmark performance and reliability by proposing Prior2Former
(P2F), the first approach for segmentation vision transformers rooted in evidential learning. P2F extends the mask vision transformer architecture by incorporating a Beta prior for computing model uncertainty in pixel-wise binary mask assignments. This design enables high-quality uncertainty estimation that effectively detects novel and OOD objects, enabling state-of-the-art anomaly instance segmentation and open-world panoptic segmentation.
Unlike most segmentation models addressing unknown classes, P2F operates without access to OOD data samples or contrastive training on \textit{void} (i.e., unlabeled) classes, making it highly applicable in real-world scenarios where such prior information is unavailable.
Additionally, P2F can be flexibly applied to anomaly instance and panoptic segmentation.
Through comprehensive experiments on the Cityscapes, COCO, SegmentMeIfYouCan, and OoDIS datasets, P2F demonstrates state-of-the-art performance across the board. %
\end{abstract}

\newcommand{\oursfull}{Prior2Former}
\newcommand{\ours}{P2F}
\newcommand{\ood}{OOD}
\newcommand{\evidential}{\text{evi}}

\section{Introduction}
Semantic understanding is essential for any autonomous agent navigating in the environment or interacting with it. State-of-the-art methods rely on comprehensive datasets that cover a data distribution of interest \cite{Cheng2020,Cheng2022} but cannot operate beyond that, limiting their applicability in real-world settings where unknown scenarios are the norm~\cite{Gasperini2023}. 
In safety-critical applications, such as mobile robotics or autonomous driving, safe operation in environments outside the training data distribution is crucial \cite{schmidt2024deep}. 
For example, an autonomous car must detect objects on the street regardless of whether their semantic class belongs to the training set and adapt its trajectory accordingly. 
Consequently, \emph{open-world} segmentation \cite{Hwang2021} and \emph{anomaly} segmentation \cite{Pinggera2016} recently emerged to study these \ood\ settings systematically.

\begin{figure}
    \vskip -0.2cm
    \centering
    \includegraphics[width=\linewidth]{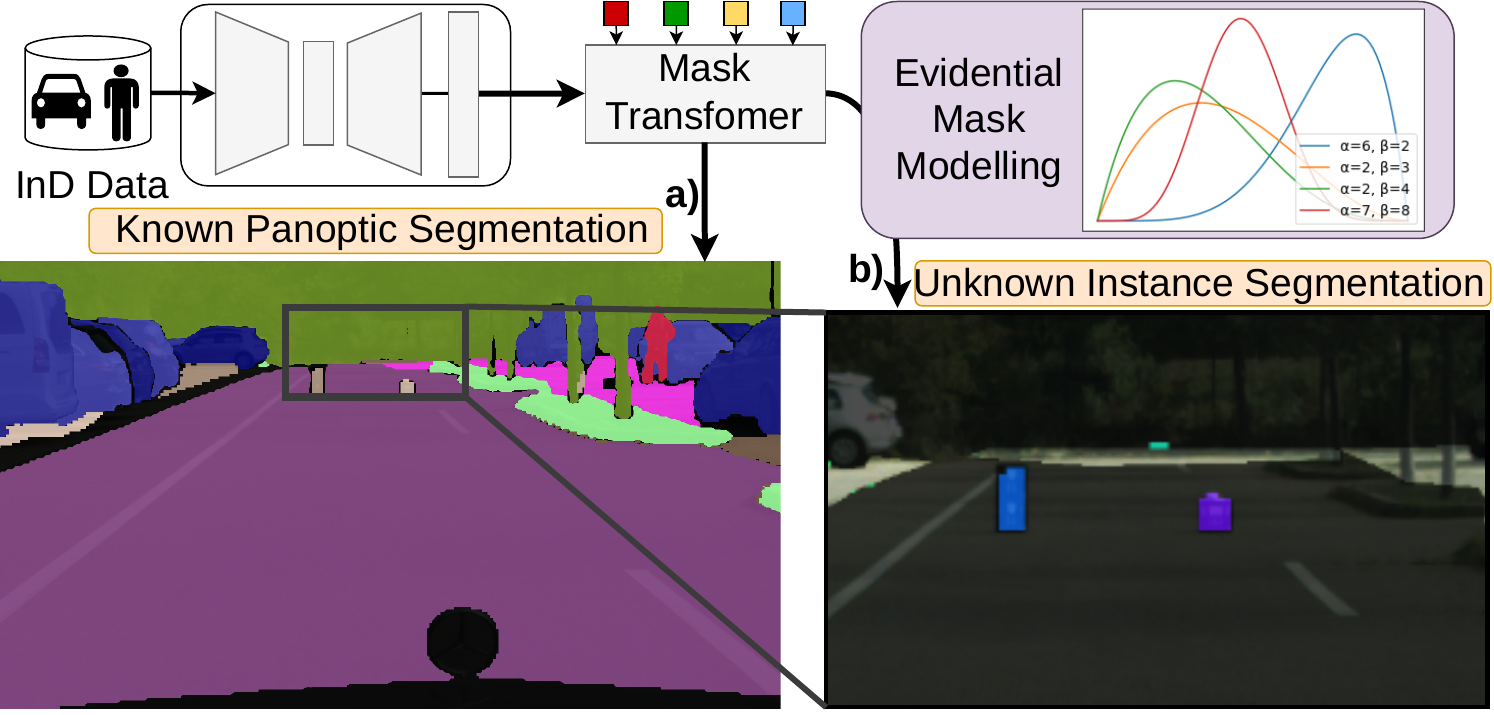}
    \vskip -0.35cm
    \caption{
    The proposed Prior2Former allows \textbf{a)} state-of-the-art panoptic segmentation performance and \textbf{b)} identifies unknown instances
    by uncertainty-based evidential learning without any knowledge beyond the standard in-domain (InD) training data.
    }
    \label{fig:TitleFig}
    \vskip -0.45cm
\end{figure}

Existing methods often approach this problem through exposure to \ood\ data~\cite{Hwang2021} or \textit{void} categories (i.e., unlabeled)~\cite{Xu2022} during training, or with powerful external models like CLIP~\cite{radford2021clip} familiar with such \ood\ data~\cite{Nekrasov2023}.  
Often used as black boxes, these external models can be large foundation models with extensive vocabularies that enable the wrapping methods to significantly expand the pool of objects they can handle~\cite{Nekrasov2023}. 
However, assuming that their knowledge covers all possible objects is unrealistic, so they fail whenever the input lies outside of their vocabulary.

As a dense task, segmentation requires a prediction for every point or pixel of the input. Some datasets annotate pixels not containing relevant elements as \textit{void}~\cite{Cordts2016}. Prior works utilize this by learning the \textit{void} areas, to which unknown classes are associated during inference~\cite{Sodano2024,Hwang2021,Xu2022}. However, they assume that the training data's \textit{void} areas are diverse enough to represent all unknowns, which is often unrealistic, limiting their ability to handle \ood\ data~\cite{Hwang2021}. Others model the generative process of unknowns~\cite{Deli2024}.

These works assume knowledge about the distribution of unknown and \ood\ objects, incorporating known OOD data into the training distribution.
However, datasets only represent a fraction of the real-world diversity, significantly restricting their applicability with unseen objects (i.e., \ood) in the real world~\cite{Gasperini2023}, despite their exceptional benchmark performance~\cite{Nekrasov2024}.
In contrast, U3HS~\cite{Gasperini2023} segments completely unseen, unknown objects without relying on external knowledge or assumptions, but it compromises performance for known classes, creating a trade-off between known and unknown capabilities.

In this paper, we overcome these issues by proposing \ours\ (\cref{fig:TitleFig}), the first vision transformer for segmentation rooted in evidential learning. It combines the Mask2Former (M2F) architecture \cite{Cheng2022} with prior networks \cite{Malinin2018} to provide reliable uncertainty estimates that enable the segmentation of unseen objects without compromising performance.

In particular, we explicitly model pixel-wise prior distributions for each binary segmentation mask query in the mask transformer architecture.
The concentration parameters of this distribution, the so-called \emph{evidence}, indicate the model's confidence and inversely correlate with its uncertainty. By learning Bayesian updates to these prior distributions, \ours\ is highly effective in various open-world applications. Notably, it does \emph{neither require any exposure to \ood\ data} nor make \emph{any other assumptions} about the unknown categories while maintaining strong performance on known classes.
We summarize our main contributions as follows:
\begin{itemize}
    \item We introduce \ours, the \emph{first} evidential mask transformer that \emph{does not require \ood\ data or knowledge} and quantifies uncertainty with negligible computational overhead.
    \item We introduce a novel loss function for training this architecture and improve existing training procedures by evidential sampling.
    \item We show that \ours\ achieves state-of-the-art performance among models that do not rely on \ood\ across the tasks of anomaly segmentation, open-world semantic and panoptic segmentation, and anomaly instance segmentation.
\end{itemize}

\section{Related Work}
\label{sec:related_work}

\textbf{Closed-World Segmentation:}
While early semantic segmentation models employ encoder-decoder CNN architectures \cite{Ronneberger2015, Chen2017, Chen2018}, recently, transformer-based models \cite{Xie2021,Cheng2021,Cheng2022} have leveraged attention to enhance performance.
Panoptic segmentation \cite{Kirillov2019} extends semantic segmentation by categorizing classes into ``\textit{stuff}'' (amorphous regions) and ``\textit{things}'' (countable objects), distinguishing also individual instances within semantic ``\textit{thing}'' classes. 
Methods such as Panoptic-DeepLab \cite{Cheng2020} provide a proposal-free approach to identify the different instances, while the masked approach of Maskformer \cite{Cheng2021} or M2F \cite{Cheng2022} directly uses masks to predict instances.
They leverage attention to query and assign masks to classes, predicting both instances and semantic \textit{stuff} classes with the same architecture. %

\textbf{Uncertainty Estimation:}
Uncertainty estimation \cite{Gawlikowski2021} is important in various domains including active learning \cite{Schmidt2020,Schmidt2023,Mukhoti2023,Schmidt2025} or \ood{} detection \cite{Schmidt2024,x_liu_y_lochman_c_zach_gen_2023} and particularly in safety-critical applications like autonomous driving~\cite{Gasperini2022}.
Approximate Bayesian \cite{Gal2015} and ensemble \cite{Lakshminarayanan2017} methods offer high-quality estimates but come at a high computational cost as they require multiple forward passes at inference time. 
Instead, sampling-free deterministic approaches estimate uncertainty in a single pass:
SNGP \cite{Liu2020a} uses spectral normalized Gaussian processes, while DDU \cite{Mukhoti2023} fits a Gaussian Mixture Model (GMM) based uncertainty estimation of DUQ \cite{Amersfoort2020}. GMMSeg \cite{Liang2022} directly incorporates a GMM into the training process.
Prior networks \cite{Malinin2018,Sensoy2018} rely on evidential learning and learn a conjugate prior to the predictive distribution from which uncertainty can be estimated. Posterior networks \cite{Charpentier2020,Charpentier2022,Stadler2021} extend this concept by learning Bayesian updates to this conjugate prior from a flow-based data density estimator. Our work builds on this principle by computing a conjugate prior for each pixel in each of the masks of the M2F architecture.

\textbf{Open-set and Anomaly Segmentation:}
Compared to closed-set standard segmentation, where only a pre-defined set of object categories is considered, in open-world or open-set settings, models encounter unknown and novel object categories that are not part of the training data \cite{Hwang2021,Gasperini2023,Schmidt2024}. 
Open-set segmentation \cite{Hwang2021} aims to classify known classes pixel-wise while identifying unknown classes
, contrasting anomaly segmentation \cite{Pinggera2016}, which only performs a binary distinction between known objects and unknowns.
Benchmarks for anomaly segmentation include Lost \& Found (L\&F) \cite{Pinggera2016}, Fishyscapes (FS) \cite{Blum2021}, and SegmentMeIfYouCan (SMIYC) \cite{Chan2021}.
Recent works combine the open-world concept with the individual instance segmentation concept and propose open-set panoptic \cite{Hwang2021} and holistic segmentation \cite{Gasperini2023} as well as the binary anomaly instance segmentation \cite{Nekrasov2024}.
Recently, \cite{sodano2024arxiv} proposed open-world panoptic segmentation with the PANIC benchmark, requiring distinguishing among unknown categories.

Open-set and anomaly segmentation approaches can be roughly categorized into three groups: 1) using additional models, 2) relying on external data and therefore making assumptions about \ood\ data, and 3) models without additional requirements.
While in the context of \ood\ detection for classification, the use of \ood\ data is explicitly prohibited \cite{Yang2022,Zhang2023}, in anomaly segmentation, using \ood\ data or external models to ensure generalization is rather common.

1) Approaches using external foundation models can leverage their knowledge from vast datasets with numerous classes. For example, UGainS \cite{Nekrasov2023} uses SAM \cite{Kirillov2023} for instance-level anomaly segmentation. This paradigm implicitly relies on the foundation model to cover all potential anomaly types, even though this is unrealistic in practice, as objects unknown to the larger model would still exist.

2) Most approaches for anomaly segmentation employ \ood\ data or \textit{void} regions that contain various elements that do not belong to any of the other labeled semantic classes.
Here, \ood\ data is used by directly training the model to assign these instances to the \textit{void} class and an additional \ood\ class \cite{Choi2023,Tian2022,Liu2020}.
Maskanomaly \cite{Ackermann2023} and RbA \cite{Nayal2022} propose mask rejection after tuning on \ood\ data. Mask2Anomaly (M2A) \cite{Rai2023} aims to learn an additional background mask from \ood\ samples.
EAM \cite{Grcic2023} generates \ood\ artifacts through patch cutting during training, while Uno \cite{Deli2024} generates synthetic \ood\ data with a normalizing flow.
The use of real and synthetic \ood\ data during training introduces a distributional prior knowledge about the anomalies that helps the model detect them.
ContMav \cite{Sodano2024} employs contrastive and objectosphere losses to \textit{void} regions for anomaly and open-world segmentation. EOPSN \cite{Hwang2021} and DDOSP \cite{Xu2022} re-identify unlabeled unknown categories that are within the training data's \textit{void} regions. %
These methods relying on \ood\ data or \textit{void} regions perform well on existing benchmarks. However, these benchmarks often satisfy the distribution assumptions that these works rely on~\cite{Hwang2021}. This makes them particularly weak whenever the \ood\ data comes from different distributions.
Therefore, \emph{in line with the holistic segmentation setting \cite{Gasperini2023}, we consider any use of (pseudo-) \ood\ data -- including post-training heuristics, external foundation models, or contrastive learning on \textit{void} regions -- to violate the open-set paradigm, as it leaks knowledge about \ood\ categories into the model.}

3) Only a few approaches are assumption-free in this regard.
SML \cite{Jung2021} achieves good unknown segmentation performance using standardized logits, while MSP \cite{Hendrycks2017} leverages softmax scores. U3HS \cite{Gasperini2023} employs a
Dirichlet Prior Network 
based on DeeplabV3+ and clustering of uncertain embeddings to segment and distinguish unknown instances. 

\emph{In this work}, we develop an uncertainty-aware transformer architecture that does not rely on assumptions about \ood\ data through evidential learning. While this paradigm has shown potential in previous segmentation models \cite{Xie2021,Cheng2021,Cheng2022,Gasperini2023}, it remains unexplored for transformer-based segmentation. Unlike prior works, this enables our models to segment any type of unknown object while performing strongly on known categories.

\begin{figure*}
    \centering
    \includegraphics[width=\textwidth]{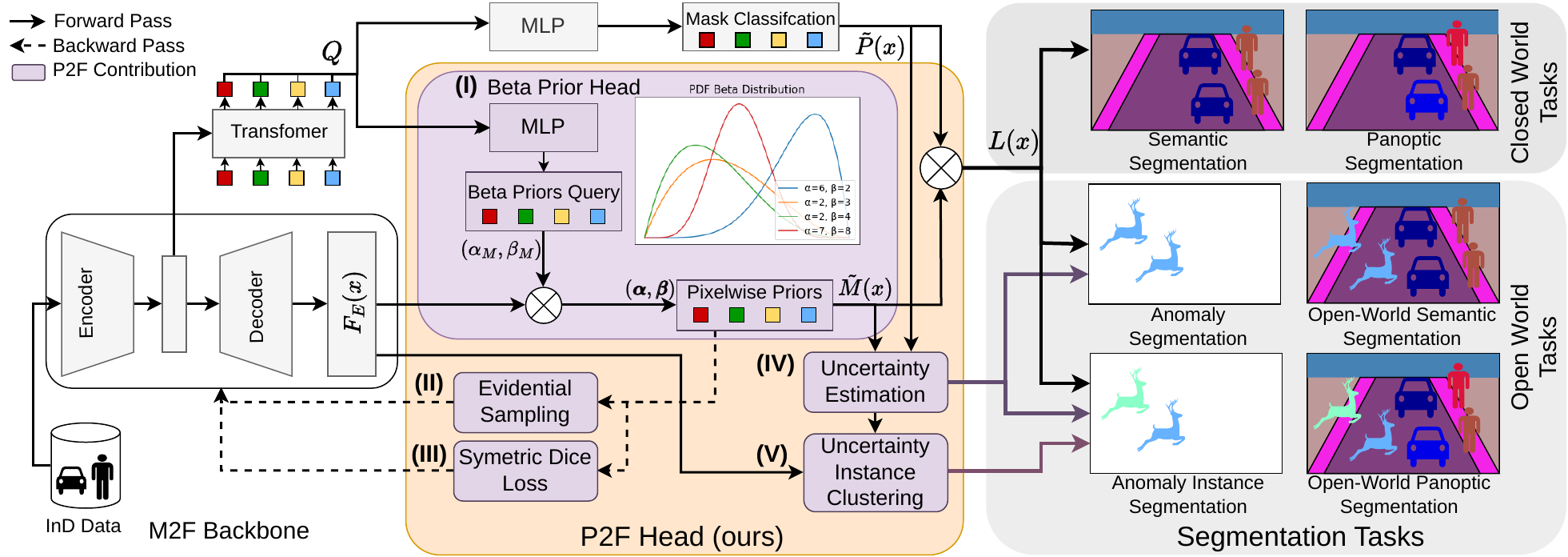}
    \vskip -0.1cm
    \caption{Overview of the Prior2Former (P2F) architecture. P2F is composed of a \textbf{beta-prior head (I)}, \textbf{evidential sampling (II)}, \textbf{symmetric Dice loss (III)}, \textbf{uncertainty estimation (IV)}, and \textbf{uncertainty instance clustering (V)}. Only standard, In-distribution data is used to train our P2F \textbf{(I)} using our novel symmetric Dice loss \textbf{(II)} and evidential sampling \textbf{(III)}. During inference, our uncertainty estimation \textbf{(IV)} detects anomalous objects, while our uncertainty instance clustering \textbf{(V)} can distinguish the instances of novel classes.
    }
    \label{fig:architecture}
    \vskip -0.1cm
\end{figure*}

\section{Preliminaries}
\label{sec:preliminaries}
We consider a dataset of RGB images $X \in \mathbb{R}^{H \times W \times 3}$ and panoptic labels $Y \in \mathbb{N}^{H \times W \times C}$.

\textbf{Mask-Based Prediction:}
In contrast to CNN architectures, which predict class logits for each pixel independently, mask-based architectures \cite{Cheng2021,Cheng2022} generate binary pixel-wise masks, which are then each classified into the $C$ segmentation labels.
First, a backbone extracts image features from which a pixel decoder computes per-pixel embeddings $F_E \in \mathbb{R}^{E \times H \times W}$. A transformer then uses masked attention to compute embeddings $F_M \in \mathbb{R}^{E \times N_M}$ for all $N_M$ masks and unnormalized per-mask class assignments $\tilde{P}_M \in \mathbb{R}^{N_M \times C}$.
The similarity between each pixel's embedding and the mask embeddings determines its (unnormalized) mask correspondence $\tilde{M} = F_M \cdot F_E \in \mathbb{R}^{H \times W \times N_M}$. 
The per-mask class assignments are normalized with softmax $P_M = \sigma_{\textbf{SM}}(\tilde{P}_M)$, while mask correspondences are normalized with a sigmoid $M = \sigma(\tilde{M})$. 
Finally, the per-pixel segmentation logits are obtained by weighting the per-mask class assignment for each pixel individually by its correspondence to each respective mask:
\begin{align}
    L(x)[h, w, c] = \sum_{i=1}^{N_M} P(x)[i, c] \cdot M(x)[h, w, i].
\end{align}
The loss is calculated using mask class assignments and per-pixel correspondences. Labels are represented as $l_X$ binary masks $Y_M \in \{0,1\}^{H \times W \times l_X}$, which correspond to class labels for each mask $Y_C \in \{1, \dots, C\}^{l_X}$.
The masks $M(x)$ are matched to the best-fitting ground-truth mask in $Y_M$ using a Hungarian matcher. The pixel-to-mask correspondence is trained using a binary cross-entropy $\mathcal{L}_{\text{BCE}}$ and a Dice loss $\mathcal{L}_{\text{Dice}}$ \cite{Sudre2017} between $M(x)$ and $Y_M$. Similarly, the per-mask class assignments are trained with a cross-entropy loss $\mathcal{L}_{\text{CE}}$ between $P(x)$ and $Y_C$.
For panoptic segmentation, the label set $Y$ consists of one binary mask for every \textit{stuff} class and one for each instance per image. %
Following M2F \cite{Cheng2022}, the loss is computed only for a subset of pixels $\mathcal{P}$.
Masks in $M(x)$ not matching any mask in $Y_M$ are ignored in the mask loss, and the mask classification loss is adjusted to assign them to an additional \emph{no-object} class.

\textbf{Evidential Learning:}
Traditional models for semantic segmentation apply a {softmax} $\sigma_{\text{SM}}$ over the class logits predicted for each pixel. Despite the ubiquity of this paradigm, recent work has found shortcomings in accurately representing model uncertainty due to a tendency toward overconfidence \cite{guo2017calibration,hein2019relu,kendall2017uncertainties}.
A promising alternative is Evidential Learning \cite{Sensoy2018}, which is grounded in the Dempster–Shafer Theory \cite{dempster1968generalization}. Rather than directly modeling a categorical distribution $p(y \mid \mathbf{x}) \sim \text{Cat}(\sigma_{\text{SM}}(L(\mathbf{x})))$ for an instance with features $\mathbf{x}$, evidential models parametrize its conjugate prior by predicting the \textit{evidence} for each possible class. 
For a categorical distribution over $C$ classes, the conjugate prior is a Dirichlet distribution parametrized by $K$ concentration parameters $\boldsymbol{\kappa} = (\kappa_1, \kappa_2, \ldots, \kappa_C)$:
\begin{equation}
    \text{Dir}(\boldsymbol{p}|\boldsymbol{\kappa})= \frac{\Gamma(\sum_{k=1}^{K}\kappa_k)}{\prod_{k=1}^{K}\Gamma(\kappa_k)}\prod_{k=1}^{K}p_k^{\kappa_k-1}
\end{equation}
Here, $\Gamma$ is the gamma function. Intuitively, the (second-order) Dirichlet is a distribution over the (first-order) categorical distributions $\boldsymbol{p}$. Predicting the conjugate prior to the predictive distribution allows a model to represent different types of uncertainty \cite{sale2023second}. Low evidence priors indicate high model uncertainty as the corresponding first-order predictive distributions are less concentrated. Conversely, high evidence priors, i.e., concentrated first-order distributions, correspond to high confidence. This enables the model to predict high-entropy first-order distributions with high confidence to represent data-inherent uncertainty. Notably, first-order distributions alone cannot differentiate between inherent and model-related uncertainty \cite{sale2023second}, a distinction crucial for downstream anomaly-related applications \cite{Mukhoti2023}.

\section{Prior2Former}
We propose to leverage the benefits of evidential learning in the context of mask transformers for segmentation to enable them to faithfully represent uncertainty by designing \ours, the first evidential mask transformer architecture. \ours\ is trained to predict a second-order prior to its binary per-pixel segmentation masks. This induces uncertainty-aware embeddings that can be used to tackle various closed-world and open-world tasks, including closed-world and open-world panoptic segmentation, anomaly segmentation, and anomaly instance segmentation. Remarkably, all of these tasks can be addressed with our same architecture and training framework, \emph{without} requiring additional data or any other assumption about potential anomalies.

Depicted in \cref{fig:architecture}, \ours{} predicts the parameters of the conjugate Beta distribution for each segmentation mask through a \textbf{beta-prior head (I)} to accurately describe model uncertainty.
Additionally, we introduce a novel \textbf{evidential} uncertainty \textbf{sampling} strategy \textbf{(II)} and propose a \textbf{symmetric Dice loss (III)} to train \ours{}. Both innovations are explicitly tailored toward Beta distributions to increase the quality of the associated mask prediction.
Lastly, we convert the predicted mask-based evidence scores into accurate \textbf{uncertainty estimates (IV)}. We finally leverage these with an \textbf{uncertainty instance clustering (V)} to distinguish between individual unknown instances.

\textbf{(I) Beta-Prior Head:} 
As described in \cref{sec:preliminaries}, the conjugate prior in classification problems is a Dirichlet distribution \(\text{Dir}(\boldsymbol{p}|\boldsymbol{\kappa}) \) over all possible categorical distributions. In binary classification, the Beta distribution is used as the conjugate prior to the predictive Bernoulli distribution.
It is parameterized by concentration parameters $\alpha, \beta > 0$:
\begin{equation}
\text{Beta}(y \mid \alpha, \beta) = \frac{\Gamma(\alpha + \beta)}{\Gamma(\alpha) \Gamma(\beta)}y^{\alpha - 1}(1-y)^{\beta - 1}.
\end{equation}
Here, $\Gamma$ is the gamma function. The concentration parameters $\alpha$ and $\beta$ can be interpreted as pseudo-counts for positive and negative observations of the corresponding Bernoulli variable. Their magnitude, therefore, serves as \textit{evidence} and can be used for uncertainty quantification.

Instead of letting the M2F architecture predict the (unnormalized) mask assignments $F_M(x)$ directly, we extend it to estimate the corresponding concentration parameters $\pmb{\alpha}_M(x), \pmb{\beta}_M(x) \in \mathbb{R}^{N_M \times E}$. Similarly to how M2F computes $\tilde{M}(x)$, we use multiplication to obtain per-pixel evidences $\pmb{\alpha}(x), \pmb{\beta}(x) \in \mathbb{R}^{H \times W \times N_M}$:
\begin{align}
    \pmb{\alpha}(x) = \pmb{\alpha}_M(x) \cdot F_E(x)
    \qquad
    \pmb{\beta}(x) = \pmb{\beta}_M(x) \cdot F_E(x)
    .
\end{align}
While the M2F architecture computes the soft pixel-to-mask assignments with an element-wise sigmoid as $M(x) = \sigma(\tilde{M}(x))$, in \ours, we propose to use the expected first-order pixel-to-mask assignment instead, as:
\begin{align}
    M(x) &= \mathbb{E}_{\mathbf{p} \sim \text{B}(\pmb{\alpha}(x), \pmb{\beta}(x))}[\mathbf{p}]
    = \frac{\pmb{\alpha}(x)}{\pmb{\alpha}(x) + \pmb{\beta}(x)}
    .
\end{align}

For each pixel, the concentration parameters $\alpha$ and $\beta$ indicate the evidence the model assigns to the positive and negative classes, respectively. Therefore, they induce an evidence-based uncertainty score:
\begin{align}
    \mathbf{U}_\evidential(x) = -(\pmb{\alpha}(x) + \pmb{\beta}(x)).
    \label{eq:evidential_uncertainty}
\end{align}
Intuitively, for a given pixel and mask, the \ours\ model assigns high uncertainty only if it does not predict any evidence for either the positive or negative class.

\textbf{(II) Evidential Sampling:}
During training, \ours{} predicts the concentration parameters of the Beta prior $\pmb{\alpha}_M, \pmb{\beta}_M$. We compute them as the similarity scores between the output of a mask query MLP and the pixel-wise representations of the decoder. By first applying a softplus function and then incrementing all values by $1$, we satisfy the positivity constraint of the concentration parameters. Consequently, the resulting Beta prior is always non-degenerate, and its expected values -- the mask probabilities $M(x)$ -- are always in the open interval $(0, 1)$.
We adapt the mask sampling procedure of M2F \cite{Cheng2022} that is outlined in \cref{sec:preliminaries} and represent the segmentation masks through $l_X$ binary ground-truth binary targets $Y_M \in \{0, 1\}^{H \times W \times l_X}$. In panoptic tasks, we use an individual mask for each instance within the image and each \textit{stuff} class. Each of the $N_M$ masks $i$ predicted by \ours\ is then matched to a mask $\hat{i}$ in $Y_M$. We refer to masks that were matched as \textit{predicted masks} and index them with $\mathcal{M}(x) \subseteq \{1, \dots, N_M\}$. For each mask $i \in \mathcal{M}(x)$, only a subset of pixels $\mathcal{P}_i(x)$ is sampled to calculate the loss with respect to the matched target mask $\hat{i}$.

We utilize the evidential uncertainty of \ours\ to improve the training efficiency through an evidence-based sampling strategy for each mask $\mathcal{P}_i(x)$. For a given budget of pixels for which the training loss is to be computed, we assign $75\%$ of it to pixels with the highest evidential uncertainty according to \cref{eq:evidential_uncertainty} and randomly select the remaining $25\%$. This matches the sampling of M2F \cite{Cheng2022} and steers the training to focus on regions with high evidential uncertainty. The corresponding loss maximizes the log-likelihood of the ground-truth masks under the predicted Beta prior:
\begin{align}
    &\mathcal{L}_\evidential(Y_M, x) = \frac{1}{\lvert \mathcal{M}(x)\rvert} \notag\\
    & \sum_{i \in \mathcal{M}(x)} \mathcal{L}_\evidential(Y_M(x)[\dots, \hat{i}], \pmb{\alpha}(x)[\dots, {i}], \pmb{\beta}(x)[\dots, {i}])
\end{align}
where $\mathcal{L}_\evidential(Y, \pmb{\alpha}, \pmb{\beta}) =$
\begin{align}
  \sum_{h, w \in \mathcal{P}_i(x)} -\log \text{Beta}(Y[h, w] \mid \pmb{\alpha}[h, w], \pmb{\beta}[h, w]).
\end{align}

\textbf{(III) Symmetric Dice Loss:}
The task of predicting binary segmentation masks is highly imbalanced. Therefore, we introduce a novel symmetric Dice loss to prevent updates that predominantly focus on $\pmb{\beta}$:
\begin{equation}
\begin{aligned}
    &\mathcal{L}_{\text{sDice}}(Y_M, x) = \frac{1}{2}\bigg( \mathcal{L}_{\text{Dice}}\left(\frac{\pmb{\alpha}(x)}{\pmb{\alpha}(x) + \pmb{\beta}(x)}, Y_M\right) \\
    &+ \mathcal{L}_{\text{Dice}}\left(\frac{\pmb{\beta}(x)}{\pmb{\alpha}(x) + \pmb{\beta}(x)}, 1 - Y_M\right) \bigg).
\end{aligned}
\end{equation}
Here, $\mathcal{L}_{\text{Dice}}$ refers to the standard Dice loss between the predicted probabilities and the true segmentation.
The total loss to train \ours{} combines the evidential and the symmetric Dice losses with the cross-entropy loss $\mathcal{L}_{\text{CE}}$ of \cref{sec:preliminaries} as:
\begin{equation}
\mathcal{L}_{P2F}=\lambda_{CE}\mathcal{L}_{CE}+\lambda_{\text{sDice}}\mathcal{L}_{\text{sDice}} +\lambda_\evidential\mathcal{L}_\evidential.
\end{equation}
We compute all losses on the same pixel sets $\mathcal{P}_i$. In contrast to other approaches, \ours\ is not explicitly trained to reject anomalous instances by assigning them to an additional \textit{void} class. 
Therefore, similarly to \cite{Gasperini2023}, we do not require any form of \ood\ data or surrogates that (implicitly) make assumptions about its distribution to train \ours.

\begin{figure*}
    \vskip -0.1cm
    \centering
    \includegraphics[width=1\linewidth]{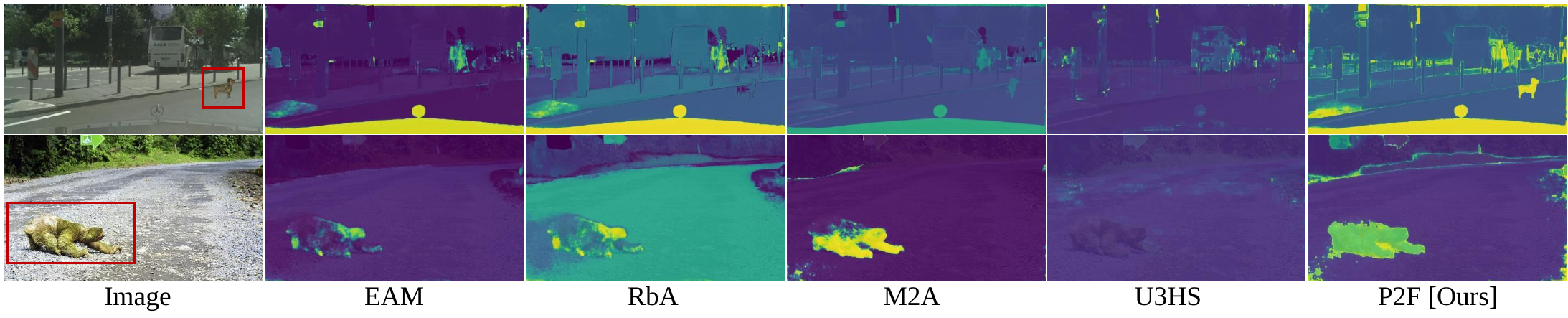}
    \vskip -0.2cm
    \caption{Qualitative comparison of uncertainty for binary anomaly segmentation on FS \cite{Blum2021} (top) and SMIYC Road Anomaly \cite{Chan2021} (bottom).}
    \vskip -0.25cm
    \label{fig:VisualSMIYC}
\end{figure*}
\begin{figure*}
    \centering
    \includegraphics[width=0.95\linewidth]{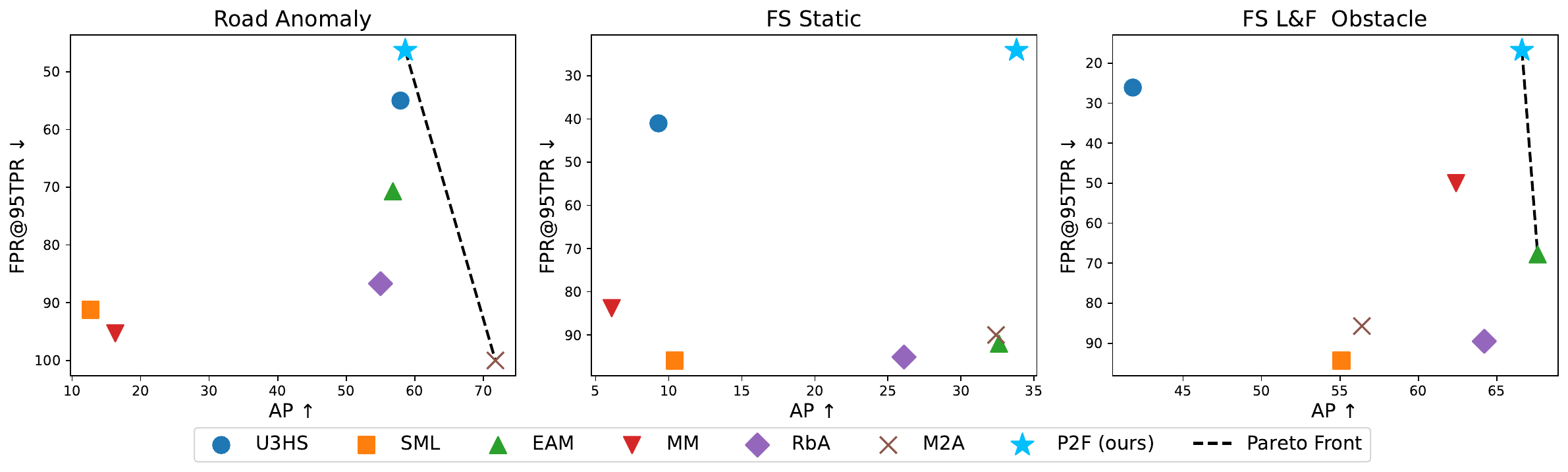}
    \vskip -0.3cm
    \caption{Anomaly Segmentation Benchmarks using the official SMIYC benchmark code \cite{Chan2021} on the validation sets for SMIYC Road Anomaly, FS Static, and the FS L\&F obstacle track. \ours{} shows the best trade-off between AP and FPR for all three benchmarks.}
    \label{fig:anomalieSegPar}
    \vskip -0.3cm
\end{figure*}

\textbf{(IV) Uncertainty Estimation:}
We now outline how the evidence-based mask uncertainty of \cref{eq:evidential_uncertainty} can be combined with the mask classification $P(x)$. First, we disregard all masks $\mathcal{M_\emptyset} \subseteq \{1, \dots, N_M\}$ that are classified as \textit{no-object}; this step is referred to as mask filtering and rejects non-primary masks from the M2F training. For each of the remaining masks, \ours\ represents its confidence in a pixel being assigned to the mask through the concentration $\pmb{\alpha}$. The confidence with which \ours\ assigns a pixel to any class is therefore represented by the highest evidence over all masks. For the corresponding mask $i^*$ (\cref{eq:maxalpha}), we obtain the probability of a pixel being assigned to this mask as the expectation under the predicted Beta prior $p^*_M$ (\cref{eq:Evalue}). We use this probability to re-weight the confidence $p^*_C$ with which the corresponding mask is classified (\cref{eq:softmaxProb}). Finally, the uncertainty $U$ associated with the segmentation of a pixel is quantified as inversely correlated with this weighted classification confidence (\cref{eq:finalunc}). Effectively, our procedure adjusts the uncertainty arising from mask classification by re-weighting it according to the evidential pixel-wise uncertainty in the predicted mask.
\begin{subequations}
    \begin{align}
        &i^*(h, w) = \argmax_{i \in \{1, \dots, N_M \} \setminus \mathcal{M}_\emptyset} \pmb{\alpha}(x)[h, w, i]
        \label{eq:maxalpha} \\
        & p_M^*(h, w) = \left(\frac{\pmb{\alpha}(x)}{\pmb{\alpha}(x) + \pmb{\beta}(x)}\right)[h, w, i^*(h, w)]
        \label{eq:Evalue} \\
        & p_C^*(h, w) = \max_c \sigma_\text{SM}(\tilde{P}_M(x)[i^*(h, w), c])
        \label{eq:softmaxProb} \\
        & U(h, w) = -p_C^*(h, w) \cdot p_M^*(h, w)
        \label{eq:finalunc} 
    \end{align}
\label{eq:uncertainty}
\end{subequations}
\indent\textbf{(V) Uncertainty Instance Clustering:}
To distinguish between different uncertain instances, we utilize the embeddings from the pixel decoder $F_E$ and our uncertainty $U$ from \cref{eq:uncertainty}.
By thresholding the uncertainty scores \cref{eq:uncertainty}, we obtain candidates for anomaly segmentation. The corresponding pixel embeddings from the pixel decoder $F_E$ are then clustered using DBSCAN \cite{DBSCAN}. 
 This supports the instance assignment as the masks instance pairing can be unreliable for \ood\ data.
Details on the threshold are given in \cref{app:experimentdDetails}.
The outlier pixels that are not assigned to any cluster are reassigned to the original semantic class and instance, respectively. In contrast to U3HS \cite{Gasperini2023}, which also employs clustering-based anomaly instance segmentation, \ours\ avoids contrastive losses, simplifying the training.

\section{Experiments}
\label{sec:Experiments}
We comprehensively evaluate \ours{} across various tasks and settings, primarily focusing on four tasks: \textbf{anomaly segmentation} and \textbf{anomaly instance segmentation} (i.e., known classes are ignored), \textbf{open-set semantic} and \textbf{open-set panoptic segmentation} (i.e., segmenting known classes, too) as in holistic segmentation \cite{Gasperini2023}.
To highlight that \ours{} addresses \emph{all tasks with the same architecture and training procedure}, we adopt the same panoptic training strategy outlined in \cref{sec:preliminaries} for all experiments. Details on the experimental settings are provided in \cref{app:experimentdDetails}.

\begin{figure*}
    \vskip -0.1cm
    \centering
    \vskip -0.2cm
    \includegraphics[width=1\linewidth]{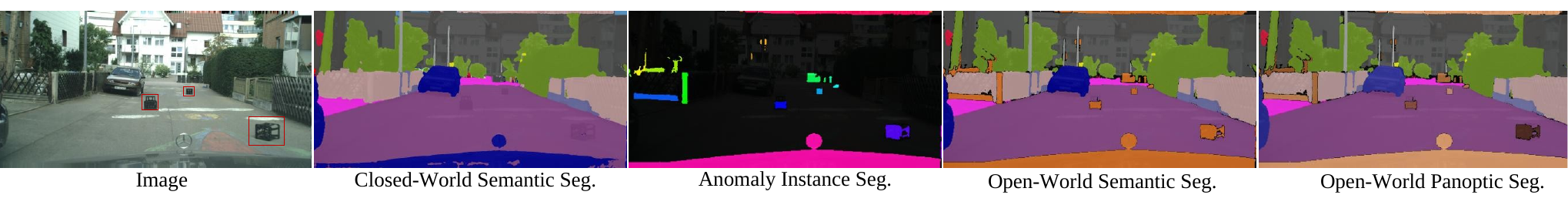}
    \caption{P2F model prediction on L\&F \cite{Pinggera2016}, part of the OoDIS \cite{Nekrasov2024}. Red boxes mark unseen instances. \ours{} effectively distinguished unknown anomalies, even the ones outside the L\&F region of interest (i.e., road), such as the trash cans at the end of the street.}
    \label{fig:AnomalyInstanceSeg}
    \vskip -0.2cm
\end{figure*}
\begin{table*}
    \centering
    \begin{tabular}{l|cc|cc|cc|cc|cc}
        \toprule
        \multicolumn{1}{c|}{} & \multicolumn{1}{c}{No} & \multicolumn{1}{c|}{No} & \multicolumn{2}{c|}{L\&F} & \multicolumn{2}{c|}{Anomaly} & \multicolumn{2}{c|}{Obstacles} & \multicolumn{2}{c}{Mean}\\
         Method & Extra Models & OOD Data & AP $\uparrow$ & AP50 $\uparrow$ & AP $\uparrow$ & AP50 $\uparrow$ &  AP $\uparrow$ & AP50 $\uparrow$ &  AP $\uparrow$ & AP50 $\uparrow$ \\
        \midrule
        UGainS \cite{Nekrasov2023} & \xmark & \xmark & 27.14 & 45.82 & 11.42 & 19.15 & 27.22 & 46.54 & 25.19 & 42.81\\
        M2A \cite{Rai2023}& \cmark & \xmark & 11.73 & 23.64 & 4.78 & 9.03 & 17.23 & 28.44 & 13.73 & 24.3 \\
        \midrule
        U3HS \cite{Gasperini2023} & \cmark & \cmark & 0.19 & 0.73 & 0 & 0 & 0.22 & 0.62 &0.19 & 0.58\\
        P2F [ours] & \cmark & \cmark & \textbf{5.60} &  \textbf{14.55}  & \textbf{0.59} & \textbf{1.25} & \textbf{2.25} & \textbf{6.88} & \textbf{3.21} & \textbf{8.85}\\
        \bottomrule
    \end{tabular}
    \vskip -0.1cm
    \caption{Comparison of anomaly instance segmentation methods on the \textbf{official OoDIS}\cite{Nekrasov2024} \textbf{benchmark}, including the three subsets L\&F, Road Anomaly and Road Obstacles. Best scores without additional requirements are in bold.}
    \vskip -0.2cm
    \label{tab:UncertaintyAnomalyInstanceBenchmark}
\end{table*}

\textbf{Anomaly Segmentation:}
First, we study the anomaly segmentation capabilities of \ours{} by comparing ours with state-of-the-art anomaly segmentation approaches.
In this binary segmentation task, anomalous objects must be identified.
We evaluate on the benchmarks L\&F \cite{Pinggera2016}, FS \cite{Blum2021}, and SMIYC \cite{Chan2021}. 
To compare fairly, we train all models with the same ResNet-50 \cite{He2016} backbone and use the same panoptic training procedure without \ood\ data. We use the official SMIYC benchmark code to compute the Average Precision (AP) and False Positive Rate (FPR) for each model.
As the driving motivation for our work is \emph{avoiding assumptions about \ood\ data}, we focus on baselines that do not rely on \ood\ data. Specifically, we employ as baselines \textbf{RbA} \cite{Nayal2022}, \textbf{EAM} \cite{Grcic2023} and \textbf{M2A} \cite{Rai2023} based on M2F and consider the general baselines \textbf{SML} \cite{Jung2021} and \textbf{MM} as mask variants of \textbf{MSP} \cite{Hendrycks2022} on M2F. While \textbf{RbA} explicitly reports results for an \ood-free version, we use the \ood-free versions of \textbf{EAM} and \textbf{M2A} presented in their corresponding ablation studies. Further, we use the prior assumptions-free U3HS \cite{Gasperini2023}, which is based on DeeplabV3+ \cite{Chen2018}. Finally, although ContMAV \cite{Sodano2024} shows promising results in anomaly segmentation, it relies on contrastive supervision on \textit{void} classes during training (which we exclude), so it is omitted here.
Implementation details are provided in \cref{app:experimentdDetails_anomseg}.

In \cref{fig:VisualSMIYC}, we compare the pixel-wise uncertainty scores predicted by different methods. While EAM, RbA, and M2A assign high uncertainties to parts of the animal only in the bottom sample, \ours{} successfully identifies the \ood\ animals in both inputs.  
In \cref{fig:anomalieSegPar}, we display the results for SMIYC Road Anomaly, FS Static, and L\&F obstacle track of FS, highlighting the trade-off between AP and FPR. \ours\ consistently balances the two metrics favorably while outperforming in both metrics for FS. For L\&F, EAM performs slightly better than \ours\ in terms of AP, but significantly worse in FPR. For SMIYC Road Anomaly, M2A realizes a good AP at the cost of a very high FPR.
These results underline the effectiveness of our uncertainty estimation in the context of anomaly segmentation. Since \ours\ detects instances, in the following, we focus on the newer and more challenging anomaly \textbf{instance} segmentation task.

\textbf{Anomaly Instance Segmentation:}
We further evaluate \ours\ on anomaly instance segmentation by submitting to the recent OoDIS benchmark \cite{Nekrasov2024}. OoDIS extends the previously used L\&F \cite{Pinggera2016}, SMIYC \cite{Chan2021} Road Anomaly, and Road Obstacles track by adding instance annotations, allowing a more complex evaluation. Because of its small and diverse objects, we consider this task as the most challenging.

In \cref{tab:UncertaintyAnomalyInstanceBenchmark}, we report the results of the official OoDIS benchmark evaluation compared to other works listed on the benchmark's official website. While M2A and UGainS show strong performances, they rely on knowledge about \ood\ data and additional external models to specifically target the anomalies that the benchmark evaluates. Together with \ours, U3HS \cite{Gasperini2023} is the only one that does not rely on any sort of \ood\ data. \ours\ outperforms it by a great margin. In \cref{fig:AnomalyInstanceSeg}, we show a prediction of \ours\ on a L\&F sample of OoDIS. Besides excellent closed-world performance, \ours{} detects multiple anomalous obstacles, including the trash bins that are not part of the Cityscapes semantic classes on which the model was trained. Additionally, panoptic L\&F results are shown in \cref{app:quali}. In \cref{app:oodsVal}, we supply the results of the OoDIS benchmark on the publicly available validation set comprising 100 L\&F samples of the FS \cite{Blum2021} split. In this setting, we again focus on U3HS \cite{Gasperini2023} as it does not require assumptions about the \ood\ data distribution. Still, \ours\ demonstrates large margins.

\begin{figure}[b]
    \vskip -0.3cm
    \centering
    \includegraphics[width=1\linewidth]{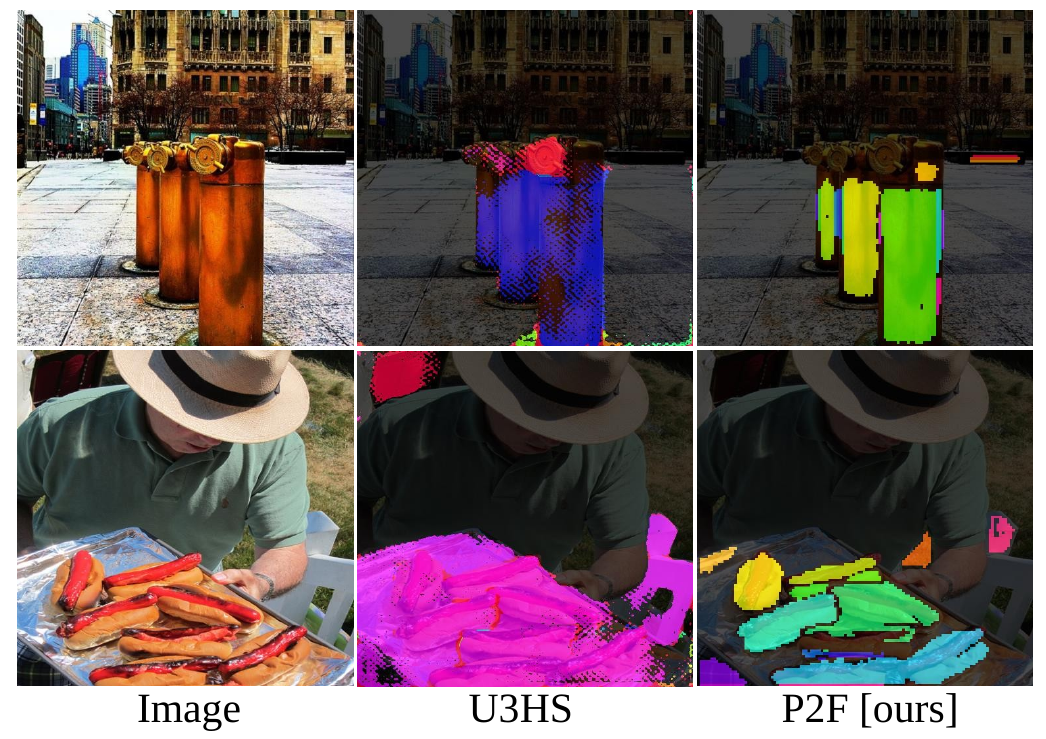}
    \vskip -0.3cm
    \caption{Comparison of anomaly instance segmentation on unseen instances on COCO \cite{Lin2014}.}
    \label{fig:OpenWorld}
    \vskip -0.1cm
\end{figure}

\textbf{Open-Set Segmentation:}
Lastly, we evaluate open-set panoptic segmentation.
We provide an evaluation of open-set semantic segmentation in \cref{app:ClosedWorldExperiments,app:SemanticSeg}. In the panoptic setting, on top of the standard, closed-set panoptic segmentation \cite{Kirillov2019}, unknown areas must be segmented as such, and each unknown instance must be distinguished by individual IDs.
Here, the number of available baselines is limited, especially when requiring models that do not rely on prior assumptions about \ood\ data, e.g., by explicitly training on the \textit{void} class. EOSPN \cite{Hwang2021} and DDOPS\cite{Xu2022} target this panoptic setting but require the \textit{void} class supervision on the unknown classes.
U3HS \cite{Gasperini2023} generalizes to all evaluated tasks and is again the only other method without any \ood\ assumptions.

\begin{table}[t]
    \centering
    \setlength{\tabcolsep}{3.2pt}
    \begin{tabular}{l|c|ccc|ccc}
    \toprule
    \multicolumn{1}{l|}{} &\multirow{2}{*}{\rotatebox[origin=c]{90}{\shortstack{No \textit{void}}}} & \multicolumn{3}{c|}{Closed-World } & \multicolumn{3}{c}{Unseen Classes} \\
     \rule{0pt}{3.5ex} Method & & PQ$\uparrow$  & SQ $\uparrow$ & RQ$\uparrow$   & PQ$\uparrow$  & SQ $\uparrow$ & RQ$\uparrow$ \\
    \midrule
    EOPSN & \xmark & 36.50 & 76.32 & 45.06 & 0.01 & 74.55 & 0.01 \\
    DDOPS & \xmark & 37.93 & 77.28 & 46.80 & 8.39 & 81.52 & 10.29 \\
    \midrule
    U3HS \cite{Gasperini2023} & \cmark & 22.03 & 69.03 & 27.92 & 9.62 & 72.84 & 13.20 \\
     P2F [ours] & \cmark & \textbf{41.34} & \textbf{78.60} & \textbf{50.79} & \textbf{10.71} & \textbf{79.96} & \textbf{13.39}  \\
    \bottomrule
    \end{tabular}
     \vskip -0.2cm
    \caption{Panoptic comparison on the COCO val.~set \cite{Lin2014} in closed-world settings and on the unseen, unknown class at full resolution.}
    \label{tab:PerformanceOpenCoco}
\end{table}
\begin{table}[t]
    \centering
    \setlength{\tabcolsep}{3.9pt}
    \begin{tabular}{l|ccc|ccc}
        \toprule
        \multicolumn{1}{c|}{} & \multicolumn{3}{c|}{Closed-World}  & \multicolumn{3}{c}{Open-World} \\
         Method & PQ$\uparrow$ & SQ $\uparrow$ & RQ$\uparrow$ & PQ $\uparrow$ & SQ $\uparrow$ & RQ$\uparrow$ \\
        \midrule
        U3HS \cite{Gasperini2023} & 46.53 &  78.87 & 58.99 & 41.21 & 79.77 & 51.67 \\
        P2F [ours] & \textbf{59.40} & \textbf{80.78} & \textbf{72.34} & \textbf{45.25} & \textbf{80.69} & \textbf{54.68} \\
        \bottomrule
    \end{tabular}
    \vskip -0.2cm
    \caption{Closed- and open-world evaluation on Cityscapes \cite{Cordts2016}.}
    \label{tab:CS_open_closed}
    \vskip -0.3cm
\end{table}
In \cref{tab:PerformanceOpenCoco}, we evaluate \ours{} on COCO using as unseen, unknown the commonly 16 left-out classes \cite{Hwang2021,Xu2022,Gasperini2023} (see \cref{ap:panopticseg}).
As evaluation metrics in open-world panoptic segmentation, we consider Panoptic Quality (PQ), Segmentation Quality (SQ), and Recognition Quality (RQ) scores for known and unseen classes, 
which do not include the set of unknown classes used for \emph{void} supervision of EOPSN and DDOPS.

While EOPSN and DDOPS leverage additional training \textit{void} classes to predict the left-out classes as \textit{void}, the uncertainty-based approaches \ours\ and U3HS can eclipse both of them in PQ and RQ, but not on SQ.
Nevertheless, our \ours\ outperforms U3HS in all three metrics, further showcasing its performance.

In \cref{fig:OpenWorld}, we show predictions on COCO \cite{Lin2014} for unseen classes of \ours{} and U3HS. For the top image, U3HS correctly identifies the hydrants as unknown, but fails to separate them into instances. In contrast, \ours{} distinguishes well single objects despite their similarity and overlap. In the lower image, U3HS provides a single instance for the whole tray (unknown, too), while \ours{} manages to segment the individual unknown hotdog instances within the tray. In general, the \ours{} manages to extract more fine-grained structures compared to U3HS. Further results and visual examples are presented in \cref{app:SemanticSeg,app:quali}. 
\begin{table}[b]
    \centering
    \setlength{\tabcolsep}{6.9pt}
    \begin{tabular}{l|ccc}
        \toprule
        Method & PQ ↑ & SQ ↑ & RQ ↑ \\
        \midrule
        U3HS \cite{Gasperini2023} & 7.94 & 64.11 & 12.37 \\
        RbA \cite{Nayal2022} + \ours\ post-proc. & 7.30 & 71.85 & 10.17 \\
        EAM \cite{Grcic2023} + \ours\ post-proc. & 8.79 & 72.09 & 12.20 \\
        M2A \cite{Rai2023} + \ours\ post-proc. & 9.91 & 73.45 & 13.49 \\
        P2F [ours] & \textbf{11.22} & \textbf{74.47} & \textbf{15.06} \\
        \bottomrule
    \end{tabular}
    \vskip -0.2cm
    \caption{Panoptic on the unseen L\&F classes in the setting of \cite{Gasperini2023} after training on Cityscapes \cite{Cordts2016}. Semantic works need \ours.}
    \label{tab:unseenLF}
\end{table}

In \cref{tab:CS_open_closed}, we report results on the Cityscapes open- and closed-world setting from \cite{Gasperini2023}, where we investigate how much the uncertainty thresholding influences the prediction in the absence of any anomalies. In \cref{tab:unseenLF}, we further evaluate on L\&F in the setting introduced by \cite{Gasperini2023}. For Cityscapes, \ours{} outperforms U3HS in all categories, but with a smaller gap in open-world settings. Especially for L\&F, \ours\ significantly improves in PQ and RQ score.
Further experiments, including the recent PANIC benchmark \cite{sodano2024arxiv}, can be found in \cref{app:furtherLandF,app:oodsVal,app:Panic}, and visual results of our mask prediction are shown in \cref{ap:maskVisu}.

\textbf{Ablation Study:}
In \cref{tab:AB_Component_final}, we study the influence of no-object mask filtering, the evidential sampling (II), and the symmetric Dice loss (III). While including all masks in \cref{eq:maxalpha} (i.e., $\mathcal{M}_\emptyset = \emptyset$) has only a minor influence on the FPR, the AP for Road Anomaly is negatively impacted by this ablation. Similarly, the symmetric Dice loss mainly boosts the FPR, and evidential sampling increases the AP. We report more detailed ablations in \cref{app:ablation}.

\begin{table}[t]
\centering
\setlength{\tabcolsep}{4.2pt}
\begin{tabular}{ccc|cc|cc}
\toprule
\multirow{3}{*}{\rotatebox[origin=c]{90}{\shortstack{Mask\\ Filtering}}} & \multirow{3}{*}{\rotatebox[origin=c]{90}{\shortstack{{\footnotesize (III)} Sym. \\ Dice}}} & \multirow{3}{*}{\rotatebox[origin=c]{90}{\shortstack{{\footnotesize (II)} Evid. \\ Sampl.}}} & & & \\
& & & \multicolumn{2}{c|}{Road Anomaly} & \multicolumn{2}{c}{FS L\&F Obstacle} \\
 &  &  & AP $\uparrow$ & FPR95 $\downarrow$  & AP $\uparrow$ & FPR95 $\downarrow$ \\
\midrule
\xmark & \cmark & \cmark & 40.5 & 46.4 & 66.4 & \textbf{16.8} \\
\xmark & \xmark & \cmark &31.7 & 62.3 & 18.5 & 28.4 \\
\cmark & \xmark & \cmark & 53.1 & 72.7 & 45.3 & 95.9 \\
\cmark & \cmark & \xmark & 47.2 & 58.9 & 23.1 & 22.2 \\
\cmark & \cmark & \cmark & \textbf{58.6} & \textbf{46.3} & \textbf{66.6} & \textbf{16.8}  \\
\bottomrule
\end{tabular}
\vskip -0.2cm
\caption{Ablation study of the main components of \ours{}.}
\label{tab:AB_Component_final}
\vskip -0.3cm
\end{table}

\textbf{Limitations:}
While our \ours\ brings remarkable improvements to various assumption-free segmentation settings of unknown objects, the gap to methods relying on \ood\ data and external models remains large (see \cref{tab:UncertaintyAnomalyInstanceBenchmark}). Furthermore, the performance degradation between closed- and open-world settings (\cref{tab:CS_open_closed}) highlights the difficulties in calibrating the uncertainty estimates and the known-unknown thresholding solely on known classes. These points can be improved with better uncertainty estimates.
\section{Conclusion}
We introduced Prior2Former (\ours), the first evidential approach for mask-based vision transformers. \ours\ quantifies uncertainty by predicting the evidence of a Beta prior to the binary segmentation masks. This informs its embeddings to enable high-quality clustering of anomalous instances. \ours\ addresses many open-set segmentation tasks (anomaly segmentation, semantic and panoptic open-world segmentation, and anomaly instance segmentation) with the same architecture and training procedure. In contrast to most prior work, \ours\ does not rely on assumptions about \ood\ data. This makes it highly applicable to real-world use cases where these assumptions are often unrealistic. Still, compared to existing assumption-free works, \ours\ delivers far superior performance, as demonstrated in our extensive experiments, effectively narrowing the gap between reliable uncertainty estimates and strong closed-set predictions.


\clearpage
\clearpage
\maketitlesupplementary

\appendix

\blankfootnote{$^*$ Equal Contribution}

\section{Experimental Details}
\label{app:experimentdDetails}
In this section, we provide details about the individual experimental setup and baselines, as well as the implementation
details. 
\subsection{Anomaly Segmentation}
\label{app:experimentdDetails_anomseg}

For evaluating anomaly segmentation, we compare \ours{} against several baselines based on M2F \cite{Cheng2022} and U3HS \cite{Gasperini2023} on the SMIYC \cite{Chan2021b} benchmark. The datasets used include SMIYC Road Anomaly \cite{Chan2021}, FS \cite{Blum2021}, and FS L\&F \cite{Blum2021}. All these benchmarks employ Cityscapes \cite{Cordts2016} as their in-distribution data source. The Cityscapes dataset consists of 19 classes — 8 categorized as ``things" and 11 as ``stuff" — captured from various cities across Germany. We employ the official SMIYC evaluation script with the addition of the FS dataset. Following the official evaluation, void-labeled regions are ignored for computing results for all three datasets. For FS L\&F, only the region of interest, typically the road in front of the vehicle, is evaluated. The official metric requires an image-shaped array with anomaly scores and dynamically selects the best-fitting threshold per image.
As \textit{avoiding prior knowledge} is a fundamental aspect of our work, we primarily focus on metric-based baselines without OOD usage. Specifically, we employ the baselines \textbf{RbA} \cite{Nayal2022}, \textbf{EAM} \cite{Grcic2023} and \textbf{M2A} \cite{Rai2023} based on M2F and implemented the general benchmarks \textbf{SML} \cite{Jung2021} and \textbf{MM} as mask variant of \textbf{MSP} \cite{Hendrycks2017} on M2F. While \textbf{RbA} reports an OOD-free version, we use the OOD-free versions of \textbf{EAM} and \textbf{M2A} presented in their ablation study. Further, we use the prior knowledge-free U3HS \cite{Gasperini2023}, which is based on DeeplabV3+ \cite{Chen2018}.

In contrast to how the respective papers evaluate these methods, EAM, RbA, and M2A are trained on the panoptic segmentation task, not on semantic segmentation.
We use the same ResNet50 (R50) \cite{He2016} backbone architecture for all methods to enable comparability between M2F-based approaches and U3HS.
Additionally, EAM, RbA, and M2A train on \ood\ data to supervise anomaly detection. We omit this to mitigate assumptions about \ood\ data to enable a fair comparison of the uncertainty metric. Therefore, all presented M2F-based baselines are primarily post-processing functions applied to the output of a standard M2F model. For M2A, a global mask attention mechanism complements the ``local" mask attention during training and inference.

For SML, we apply the maximum logit standardization to the logits $L(x)$ of the M2F model introduced in \cref{sec:preliminaries}:
\begin{align}
    &L(x)_c = \sum_{i=1}^{N_m} p_i(c) \cdot m_i[h, w],
\end{align}
Let $\mathbb{T}$ be the set of all training images and let $L(x)_c \in \mathbb{R}^{H \times W}$ be the pixel-wise logits of class $c$. We compute the mean and standard deviation $\mu, \sigma \in \mathbb{R}^K$  of these maximum logits over all classes:
\begin{equation}
\begin{aligned}
&\mu_c = \\
&\text{MEAN}(\{L_{\hat{c}(x)}[h,w]  \mid 1 \leq h \leq H, 1 \leq w \leq W, x \in \mathcal{X}_\text{train}\})
\end{aligned}
\end{equation}
\begin{equation}
\begin{aligned}
&\sigma_c = \\
&\text{STD}(\{L_{\hat{c}(x)}[h,w] \mid 1 \leq h \leq H, 1 \leq w \leq W, x \in \mathcal{X}_\text{train}\})
\end{aligned}
\end{equation}
where 
\[
\hat{c}(x) = \underset{c \in \{1,\dots,K\}}{\text{arg max}}\ \  L_c(x)[h,w].
\]
Hence,
\[
U^{\text{SML}}[h,w] = -\frac{L_{\hat{c}}[h,w] - \mu_{\hat{c}}}{\sigma_{\hat{c}}}.
\]
 To obtain the SML uncertainty $U^{\text{SML}}[h,w]$ for a pixel at position $h,w$ of image $x$, we standardize the maximum logit $L^{(x)}_{\hat{c}}[h,w]$ using the mean and standard deviation computed from all pixels $[h,w]$ over all images $x$ in the training set, where the class $\hat{c}$ has the greatest logit.

The Maximum Mask (MM) baseline presents a mask-based variant of MSP \cite{Hendrycks2017}:
\[
    U^{\text{MM}}[h,w] = - \max_{i\in\{1,\dots,N_m\}} m_i[h, w].
\]
For MSP \cite{Hendrycks2017}, the maximum softmax probability of the current pixel is taken as the confidence score, whereas for MM, the maximum sigmoid probability over all masks serves as the confidence score.
The uncertainty estimates according to EAM \cite{Grcic2023}, RbA \cite{Nayal2022}, and M2A \cite{Rai2023} follow the corresponding papers:
\begin{align}
    &U^{(\text{EAM})}[h,w] = -\sum_{i=1}^{N_m} m_i[h,w] \cdot \left (\max_{c=1,...,K}(p_i(c) \right ),\\
    &U^{(\text{RBA})}[h,w] = -\sum_{c=1}^{K} \text{tanh}(L_c[h,w]),\\
    &U^{(\text{M2A})}[h,w] = \left ( 1-\max_{c=1,...,K} L_c[h,w] \right ) \cdot R_M[h,w], 
\end{align}
where the sigmoid function is applied elements-wise and $R_M$ is the mask filter as presented in M2A \cite{Rai2023}, such that the uncertainty $U^{(M2A)}[h,w]$ is set to zero if there is no mask for pixel $[h,w]$ where the masks score $m_i[h,w]>0.5$ and the predicted class is ``road" or a ``thing" class and the softmax confidence is greater than 0.95. We evaluate M2A by only using the pixel-wise filter $m_i[h,w]>0.5$ since all other models do not include class-specific information for the detection of anomalies. Note that for the M2A uncertainty, the logits are obtained from a model trained with global mask attention. 

Since the SMIYC metric expects one anomaly score per pixel on a given image, the logical ``and" between thresholded distances and anomaly scores, as implemented by U3HS \cite{Gasperini2023}, can not be used for the evaluation metric. Hence, for U3HS, the Dirichlet strength from the semantic head is used as an anomaly score.

\subsection{Anomaly Instance Segmentation}

The task of Anomaly Instance Segmentation is evaluated using the official OoDIS benchmark code \cite{Nekrasov2024}. This benchmark assesses performance on three datasets: an unknown split of L\&F \cite{Pinggera2016}, as well as the test splits of the SMIYC RoadAnomaly21 \cite{Chan2021} and RoadObstacles21 \cite{Chan2021}. 
The evaluation requires binary images containing the recognized anomalies and a confidence score for each binary image. An anomaly instance prediction with an Intersection over Union (IoU) greater than 0.5 and the highest confidence score is considered a positive prediction, contributing to the true positive and false positive rates. All other predictions, not counted as valid predictions for another anomaly, are considered false positives.

For submission and evaluation on the publicly available validation set containing only L\&F \cite{Pinggera2016} data, \ours{}, and U3HS \cite{Gasperini2023} are trained on the Cityscapes dataset.

\subsection{Closed-World and Open-World Panoptic Segmentation}
\label{ap:panopticseg}

Here, we detail the experimental setup for computing the PQ and mIoU metrics on the two datasets COCO \cite{Lin2014} and BDD100k \cite{Yu2020} for \ours{} and U3HS \cite{Gasperini2023}.

\textbf{COCO:} For the COCO \cite{Lin2014} dataset, we preprocess the data by excluding all images containing any of the 20\% (i.e., 16) least frequent classes in the training set, specifically: \emph{baseball bat, bear, fire hydrant, frisbee, hairdryer, hot dog, keyboard, microwave, mouse, parking meter, refrigerator, scissors, snowboard, stop sign, toaster, and toothbrush}, which aligns with \cite{Gasperini2023,Xu2022}. 
The images in the validation set containing any held-out classes are placed into a separate set, termed the open-world validation set, while the remaining images form the closed-world validation set. The semantic labels of the known classes are retained, while the semantic labels of the held-out classes are merged into a single \ood\ class.
We then evaluate the mIoU and PQ of the open-world validation set, treating all anomalies as belonging to the \ood\ class.
We use COCO panopticapi\footnote{\urlstyle{same}\url{https://github.com/COCOdataset/panopticapi}} and torchmetrics\footnote{\urlstyle{same}\url{https://lightning.ai/docs/torchmetrics/stable/}} for PQ and mIoU evaluation, respectively.
Following the training scheme presented in M2F \cite{Rai2023} for COCO, we employ a random resize crop augmentation strategy during training, which maintains the aspect ratio of the input images.
During validation, we maintain the original size of the COCO images. 
This approach differs from the scheme used in U3HS, where images are resized for both training and evaluation.

\textbf{BDD100k:} For the BDD100k \cite{Yu2020} dataset, we follow the evaluation and training procedure employed for COCO.
For the dataset split, we use the proposed class settings of BDD anomaly \cite{Hendrycks2022} and exclude the classes motorcycle and bicycle. In contrast to \cite{Hendrycks2022}, we use the panoptic labels instead of the semantic labels to enable a panoptic evaluation. Note that besides the instance information, the panoptic labels additionally have an increased number of classes compared to the semantic labels, which increases complexity.
We construct the closed-world training and validation sets and the open-world validation set similarly, but evaluate all scores based on the respective torchmetrics implementation.
We restrict the evaluation of mIoU and PQ only on instances that cover more than 2,500 pixels. This is because the depth values in the dataset show significant variability, resulting in many small anomalies that cover only a few pixels. Including these small anomalies in the evaluation does not meaningfully contribute to understanding the models' ability to recognize the held-out classes.

\textbf{Cityscapes:} For open and world segmentation, we follow the setup described by \cite{Gasperini2023} and use the standard training setup as suggested by M2F for all M2F-based baselines, including \ours{}. For U3HS \cite{Gasperini2023}, we use the setup described in their paper and use the respective evaluation settings. 
For the Lost \& Found evaluation, we use a resized version as done by \cite{Gasperini2023} and a full-size version to maintain a fair evaluation. 

\subsection{Implementation Details}

We train \ours{} and M2F on \textbf{Cityscapes} using a series of augmentations. Specifically, we apply a random zoom crop with a crop size of $(1024, 512)$ and a zoom range of $(1.0, 2.0)$. Additionally, we utilize a color jitter augmentation with a brightness delta of $32$, a hue delta of $18$, contrast adjustments in the range of $(0.5, 1.5)$, and saturation changes within $(0.5, 1.5)$. We also include a random horizontal flip with a probability of $0.5$. The color jittering is necessary because we use the models trained on Cityscapes for evaluation on SMIYC and Oodis. Without this augmentation, different lighting conditions, especially of the sky, could result in high anomaly scores.

We train the models with a batch size of $16$, a learning rate of $0.0001$, and a learning rate of $0.00001$ for the backbone, using the AdamW optimizer \cite{Loshchilov2019} with a weight decay of $0.05$. A polynomial learning rate scheduler with a power of $0.9$ is employed. Gradients are clipped at $0.01$ according to the L2 norm. The training is conducted for $450$ epochs, which corresponds to approximately $90,000$ iterations.
Hence, we follow the training as suggested for M2F \cite{Cheng2022} on Cityscapes, with the only difference of adding the color jittering for the more robust uncertainty estimation.

For the \textbf{COCO} dataset, we employ a random zoom crop that keeps the aspect ratio fixed with a zoom range of $(0.1, 2.0)$ and a crop size of $(512, 512)$. No color jittering is used. The optimizer and scheduler configurations, random horizontal flip, and gradient clipping remain the same, but the batch size is increased to $32$, and the training is performed over $50$ epochs.
In summary, we use a training scheme closely matching the M2F \cite{Cheng2022} configuration for COCO. However, we changed the crop size from $(1024, 1024)$ to $(512, 512)$ and the batch size from $16$ to $32$ for known settings, which improved the convergence speed.

For the \textbf{BDD} dataset, M2F \cite{Cheng2022} does not provide a configuration. Hence, we primarily follow the training on Cityscapes as a related dataset. We scale the images within the range of $(1.0, 2.0)$ and crop both dimensions by half to a size of $(640, 360)$, similar to the cropping applied in Cityscapes. No color jittering is used. All other parameters are consistent with those used for Cityscapes.

Across \textbf{all} datasets, we train using a no-object loss coefficient of $0.1$, a class loss weight $\lambda_{\text{cls}} = 2.0$, a symmetric dice loss weight $\lambda_{\text{sDice}} = 5.0$, and an evidential loss weight $\lambda_{\text{evi}} = 0.1$ for \ours{}. The evidential loss weight is tuned to ensure that the loss values of the symmetric dice and evidential loss have similar absolute values. 
We use $200$ object queries for both the baselines and \ours{}, which is necessary to generate a sufficient number of valid mask predictions on the SMIYC Road Anomaly and SMIYC Road Obstacle datasets. All other hyperparameters for model building and training remain consistent with those used in the official M2F repository \footnote{\urlstyle{same}\url{https://github.com/facebookresearch/Mask2Former}}.

For \textbf{postprocessing} the predictions of \ours{}, we set the object-mask-threshold to $0.5$, compared to $0.8$ in the original M2F model. This threshold on the mask prediction probability determines if an object is present in the panoptic prediction. However, since the beta prior predictions of \ours{} are more restrictive, we choose a lower threshold value.

To create the anomaly instance segmentation, we introduce a threshold $t$ to the uncertainty estimates of \ours{}. The corresponding feature vectors of the pixel embedding $F_E$ are then clustered using DBSCAN \cite{DBSCAN} with parameters $\text{eps}$ and $\text{min-samples}$. The mask correspondence is determined using a scalar product between the predicted mask features and the pixel embedding. Hence, it is natural to use one minus the cosine similarity instead of an Euclidean distance for the DBSCAN, as it is the normalized scalar product of the two vectors.
For the submission on OoDIS, the uncertainty threshold $t$ is set to $2$ times the standard deviation away from the mean uncertainty on the training set $t=-0.6$ for the L\&F split.
The overall uncertainty on SMIYC Road Anomaly and SMIYC Road Obstacle is greater since the images cover uncommon scenarios and conditions.
Hence, we increase the threshold to $3.5$ times the standard deviation, i.e., $t=-0.4$.
For COCO and BDD, we follow the same approach of determining the threshold and, therefore, set $t = -0.55$ and $t = -0.6$.
The parameter $\text{eps}$ strongly influences the granularity of the clustering algorithms. Evaluating the embedding space, we use $0.04$ in all experiments. Except for COCO, we set $\text{eps} = 0.1$ due to the wider variety of semantic classes and instances. The $\text{min-samples}$ parameter is set to $17$, however, clustering the embedding is robust against changes in this parameter, with values ranging from $10$ to $23$ being effective.

\section{Additional Experiments}
\label{app:AdditionalExperiments}
Besides the experiments provided in \cref{sec:Experiments}, we report additional experiments on closed-world segmentation, open-world semantics segmentation, the OoDIS validation set, as well as a further ablation study of our uncertainty metric.

\subsection{Closed World Segmentation}
\label{app:ClosedWorldExperiments}
To compare \ours{} in closed-world segmentation, we compare it to the vanilla M2F and a naive Dirichlet Prior network (DPN) \cite{Malinin2018} for M2F for Cityscapes. We follow the training settings provided in \cite{Cheng2022} and use the Cityscapes \cite{Cordts2016} script for evaluation. Besides the mIoU, we also report the category-wise IoU (cIoU).
In \cref{tab:PerformanceCityscapes}, it can be seen that P2F performs similarly to M2F. This contrasts with the massive performance drop of the DPN head.
\begin{table}
    \setlength{\tabcolsep}{6.8pt}
    \centering
    \begin{tabular}{l|cccccc}
    \toprule
    Model & mIoU ↑ & cIoU ↑ & PQ ↑& SQ ↑& RQ ↑\\
    \midrule
    M2F & 77.3 & 90.2 & 60.29 & 81.28 & 73.15 \\
    \midrule
    M2F DPN & 64.5 & 88.4 &50.29 & 67.20& 61.70 \\ %
    P2F [ours] & \textbf{77.0} & \textbf{89.1} & \textbf{59.41} & \textbf{80.74} & \textbf{72.34} \\
    \bottomrule
    \end{tabular}
    \caption{Comparison of M2F with different evidential heads on Cityscapes on Panoptic Segmentation. Best evidential heads are marked in bold.}
    \label{tab:PerformanceCityscapes}
\end{table}

\subsection{Open-World and Closed-World Segmentation}
\label{app:SemanticSeg}

Besides the results reported on panoptic segmentation in \cref{sec:Experiments}, we study open- and closed-world semantic segmentation. In \cref{tab:SematicPerformance}, we show the mIoU results for BDD100k \cite{Yu2020} anomaly and COCO \cite{Lin2014} with the left-out classes listed in \cref{ap:panopticseg}. Like in panoptic segmentation, \ours{} ranks the highest in all settings. 

\begin{table}
    \vskip -0.1cm
    \centering
    \setlength{\tabcolsep}{9pt}
    \begin{tabular}{l|cc|cc}
    \toprule
    \multicolumn{1}{c|}{} & \multicolumn{2}{c|}{BDD Anomaly} & \multicolumn{2}{c}{COCO} \\
    Method & Closed & Open & Closed & Open \\
    \midrule
    U3HS \cite{Gasperini2023} & 29.16 & 16.32 & 33.19 & 22.77 \\
    P2F [ours] & \textbf{35.73} & \textbf{29.12} & \textbf{46.00} & \textbf{33.56}\\
    \bottomrule
    \end{tabular}
    \vskip -0.1cm
    \caption{Open- and closed-world semantic segmentation comparison using \textbf{mIoU} metric for BDD Anomaly on COCO.}
    \label{tab:SematicPerformance}
    \vskip -0.1cm
\end{table}

In \cref{tab:BDD_open_closed} we report the panoptic quality metric of U3HS and \ours{} on BDD100k \cite{Yu2020} anomaly dataset. U3HS shows a strong closed-world performance in PQ and RQ. However, \ours{} achieves the highest scores in the open-world setting as well as for SQ in the closed-world setting.
\begin{table}
    \centering
    \setlength{\tabcolsep}{3.5pt}
    \begin{tabular}{l|ccc|ccc}
        \toprule
        \multicolumn{1}{c|}{} & \multicolumn{3}{c|}{Closed-W.}  & \multicolumn{3}{c}{Open-W.} \\
         Method & PQ$\uparrow$ & SQ $\uparrow$ & RQ$\uparrow$ & PQ $\uparrow$ & SQ $\uparrow$ & RQ$\uparrow$ \\
        \midrule
        U3HS \cite{Gasperini2023} & \textbf{36.82} & 80.93  & 16.32 &  17.81 & 75.59 & 23.63 \\
        P2F [ours] & 32.67 & \textbf{81.68} & \textbf{29.0} & \textbf{29.10} & \textbf{79.43} & \textbf{36.66} \\
        \bottomrule
    \end{tabular}
    \caption{Closed and open-world evaluation on the BDD100k anomaly dataset \cite{Yu2020}.}
    \label{tab:BDD_open_closed}
\end{table}

\subsection{Further Experiments on L\&F}
\label{app:furtherLandF}
To further compare using the setting of unseen L\&F data as introduced by \cite{Gasperini2023}, we compare the reduced size evaluation with the baselines reported by \cite{Gasperini2023} and an M2F confidence uncertainty. It can be seen that the additional baselines struggle with this task. The confidence baseline of M2F shows a surprisingly strong performance.
\begin{table}[h]
    \centering
    \begin{tabular}{l l c c c}
        \toprule
        Method & Assumptions & PQ ↑& SQ ↑ & RQ ↑\\
        \midrule
        EOPSN \cite{Hwang2021} & data, void & 0$^*$ & 0$^*$ & 0$^*$ \\
        OSIS \cite{OSIS} & data, void & 1.45 & 65.11 & 2.23 \\
        U3HS \cite{Gasperini2023} & none & 7.94 & 64.24 & 12.37 \\
        M2A* \cite{Rai2023}& none & 9.91 & 73.45 & 13.49 \\
        M2F* & none & 9.02 & \textbf{75.34} & 11.98 \\
        P2F [ours] & none & \textbf{11.22} & 74.47 & \textbf{15.06} \\
        \bottomrule
    \end{tabular}
    \caption{Results on the Lost\&Found (unseen) dataset with the settings of \cite{Gasperini2023}. \cite{Hwang2021} and \cite{OSIS}, are taken from \cite{Gasperini2023}. * Uses \ours\ postprocessing.}
    \label{tab:lost_found}
\end{table}

In addition to the setting reported by \cite{Gasperini2023}, we report the performance of \ours{} in comparison to other masked-based uncertainty methods in \cref{tab:anomaly_metrics} using the full resolution of the L\&F dataset. We train M2A without OOD data. We further report M2F using a classical confidence measure as uncertainty, which diverges for this task. We also report the uncertainty measures of RbA \cite{Nayal2022} and EAM \cite{Grcic2023}.
These results show that all these methods profit heavily from the increased resolution. Nevertheless \ours{} maintains the highest scores for PQ and RQ.
\begin{table}[h]
    \centering
    \begin{tabular}{lccc}
        \toprule
        Method & PQ ↑ & SQ ↑ & RQ ↑ \\
        \midrule
        M2A \cite{Rai2023} + \ours\ post-proc. & 21.06 & \textbf{73.14} & 28.79 \\
        M2F & 0.00 & 0.00 & 0.00 \\
        EAM \cite{Grcic2023} + \ours\ post-proc. & 18.39 & 71.35 & 25.78 \\
        RbA \cite{Nayal2022} + \ours\ post-proc. & 15.54 & 70.65 & 21.99 \\
        P2F [ours] & \textbf{22.07} & 69.54 & \textbf{31.73}\\
        \bottomrule
    \end{tabular}
    \caption{Unseen L\&F performance metrics on full-size resolution. Semantic approaches require \ours\ post-processing.}
    \label{tab:anomaly_metrics}
\end{table}

\begin{figure*}
    \centering
    \includegraphics[width=1\linewidth]{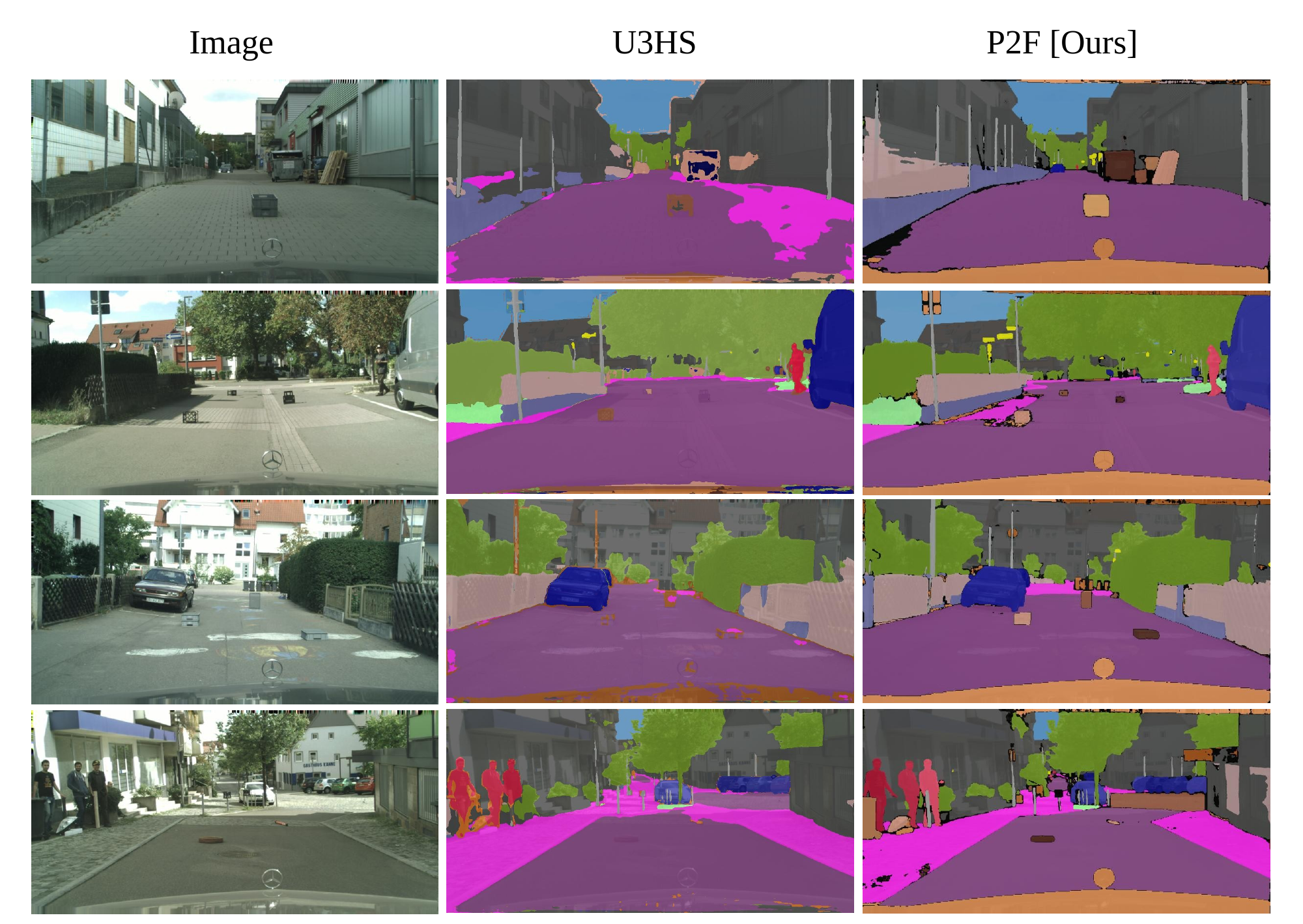}
    \caption{Open-World panoptic segmentation on L\&F test set after training on Cityscapes \cite{Cordts2016}.}
    \label{fig:LandF_open_world_panoptic}
\end{figure*}

\subsection{OoDIS Validation Set}
\label{app:oodsVal}
In addition to the official OoDIS \cite{Nekrasov2024} benchmark scores, we present results on the validation set of OoDIS comprising 100 L\&F \cite{Pinggera2016} instance anomaly labels. We report the scores for the \ood-free and extra-model-free U3HS and \ours{} in \cref{tab:UncertaintyAnomalyInstanceVal}. For U3HS, we experimented with 4 different uncertainty thresholds and reported the best. The results of both models are slightly better compared to the benchmark evaluation on a larger L\&F, while their ranking remains unchanged.

\begin{table}[tbh]
    \centering
    \setlength{\tabcolsep}{3pt}
    \begin{tabular}{@{}ll|cccc@{}}
        \toprule
        \multicolumn{2}{l|}{} & \multicolumn{1}{c}{No Aux.} & \multicolumn{1}{c}{No OOD} & \multicolumn{2}{c}{} \\
         Backbone & Model & Models & Data & AP $\uparrow$ & AP50 $\uparrow$ \\
        \midrule
        R50 &U3HS~\cite{Gasperini2023} & \cmark & \cmark & 0.61 & 2.04\\
        R50 &P2F [ours] & \cmark & \cmark & \textbf{8.17} & \textbf{16.13} \\
        \bottomrule
    \end{tabular}
    \caption{Comparison of different Anomaly Segmentation Methods on the validation set of OoDIS.
    }
    \label{tab:UncertaintyAnomalyInstanceVal}
    \vskip -0.1cm
\end{table}

\subsection{Panic Open-Set Panoptic Segmentation}
\label{app:Panic}
For additional evaluation of open-set panoptic segmentation, we utilized the novel PANIC benchmark \cite{sodano2024arxiv}. The benchmark contains images of different resolutions from different cities in Germany and evaluates the PQ, SQ, and RQ scores of open-set objects. In \cref{tab:Panic}, we show the official benchmark results of the open-set panoptic segmentation task. It can be seen that \ours{} significantly leads the benchmark introduced with the Con2Mav approach, with \ours\ delivering nearly double scores on PQ and RQ.

\begin{table}[b]
    \centering
    \begin{tabular}{l|ccc}
        \toprule
        Method & PQ ↑ & SQ ↑ & RQ ↑ \\
        \midrule
        Con2Mav \cite{sodano2024arxiv} & 21.6 & 72.4 & 28.4 \\
        P2F [ours] & \textbf{52.9} & \textbf{87.1} & \textbf{56.7} \\
        \bottomrule
    \end{tabular}
    \vskip -0.2cm
    \caption{Leaderboard of the PANIC \cite{sodano2024arxiv} Open-set Panoptic Segmentation benchmark. Results from the test set.
    }
    \label{tab:Panic}
\end{table}

\begin{figure*}
    \centering
    \includegraphics[width=1\linewidth]{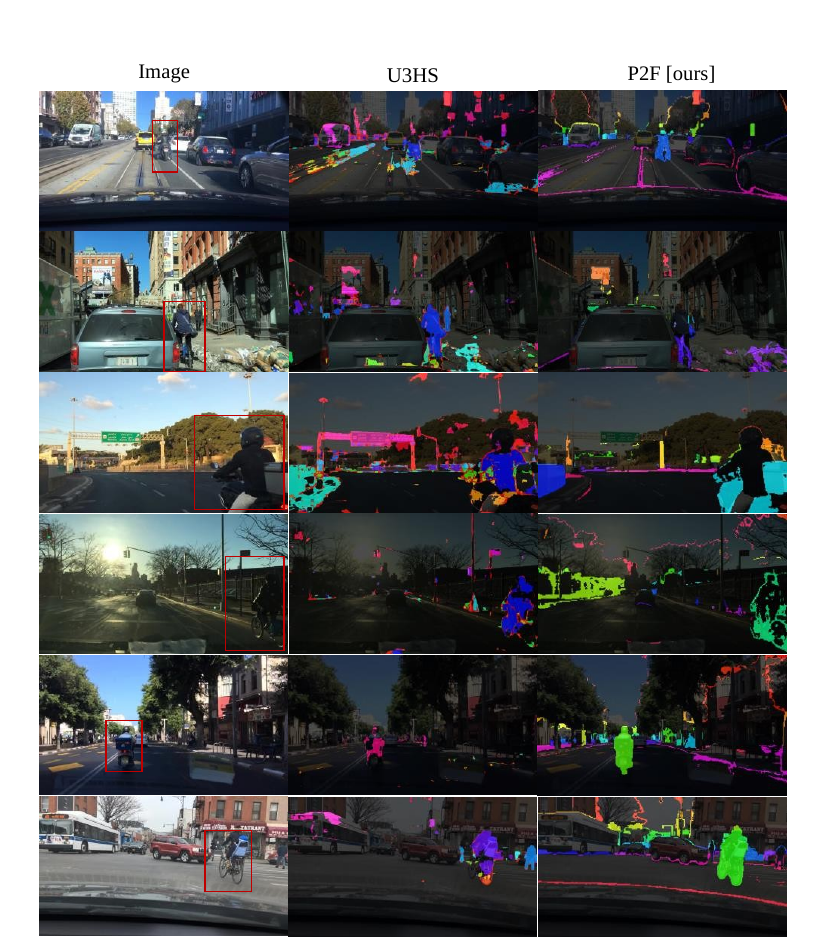}
    \caption{Visual comparison of anomaly instance segmentation on held-out classes on BDD \cite{Yu2020}, marked with a red box on the input image (left). The high diversity and unbalanced class distribution seem to confuse the DPN-based U3HS \cite{Gasperini2023}. Nonetheless, U3HS showed a strong segmentation quality for the rare classes, like the traffic sign poles. \ours{}, managed to detect the unknown motorcycle or bicycles more precisely, given its less class imbalance affected Beta's prior approach.}
    \label{fig:VisualBDD}
\end{figure*}

\begin{figure*}
    \centering
    \includegraphics[width=1\linewidth]{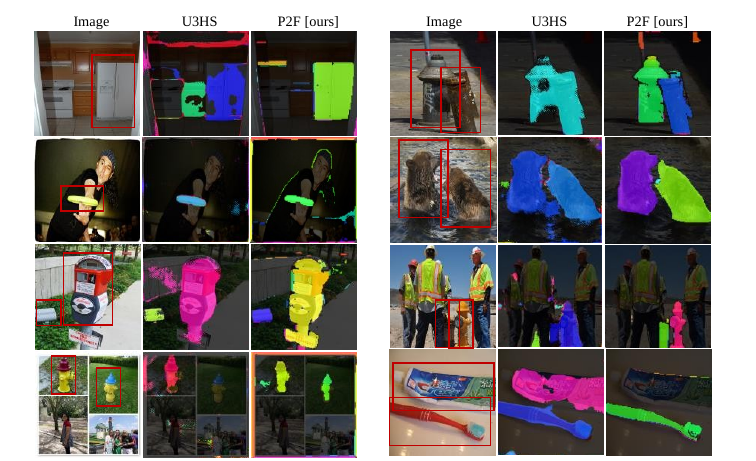}
    \caption{Visual comparison of anomaly instance segmentation on held-out classes on COCO \cite{Lin2014}, marked with a red box on the input image (left). It can be seen that U3HS shows a strong performance for individual and larger objects, but can get confused with related objects, like different kitchen objects in the top left. Additionally, it suppresses the combined uncertainty and distance approach uncertainty at the object border. With the direct mask supervision and Beta prior \ours{} provides a more robust separation of different instances, while improving the segmentation quality.}
    \label{fig:VisualCOCO}
\end{figure*}

\subsection{Uncertainty Ablation Study}
\label{app:ablation}
In this section, we conduct further ablation studies in addition to the reported study in \cref{sec:Experiments}. 
In \cref{tab:AB_Component_final_unc}, we compare the predictive uncertainty suggested by M2F \cite{Cheng2022} with the uncertainty of \ours{}. It can be seen that the \ours{} uncertainty is more effective on AP for the Road Anomaly split of SMIYC and provides a major improvement in the FPR of both datasets. This underlines the benefit of the evidential mask selection in our uncertainty.
Overall, our \ours{} shows strong uncertainty statistics. Nevertheless, for minor semantics shifts like L\&F, the mask filtering seems to be less important.

\begin{table}[]
\centering
\setlength{\tabcolsep}{2.5pt}
\begin{tabular}{@{}l|cc|cc@{}}
\toprule
\multicolumn{1}{c|}{}& \multicolumn{2}{c|}{Road Anomaly}  & \multicolumn{2}{c}{FS L\&F Obstacle} \\
Method & AP $\uparrow$ & FPR95 $\downarrow$ & AP $\uparrow$ & FPR95 $\downarrow$ \\
\midrule
Prediction Uncertainty & 48.6 & 62.2 & 61.2 & 88.4   \\
\ours{}\ uncertainty [ours] & \textbf{58.6} & \textbf{46.3} & \textbf{66.6} & \textbf{16.8}  \\
\bottomrule
\end{tabular}
\caption{Uncertainty comparison of \ours\ using the classical prediction uncertainty and \ours{} uncertainty definition.}
\label{tab:AB_Component_final_unc}
\end{table}

In \cref{tab:naive_uncertainties}, we compare our \ours{} uncertainty against three further uncertainty variants for anomaly segmentation and open-world panoptic segmentation. $\sigma$-unc. applies our combined uncertainty concept with mask matching according to \cref{eq:maxalpha} with $\sigma$ of the vanilla M2F model. Naive Beta describes the vanilla evidential uncertainty according to \cref{eq:evidential_uncertainty} and M2F* the vanilla M2F uncertainty with our postprocessing. It can be seen that our combination concept of mask selection and evidential uncertainty outperforms our concept on M2F and the vanilla evidential uncertainty. 

\begin{table}[h]
\vskip -0.2cm
\centering
\begin{tabular}{l|cc|ccc}
\toprule
& \multicolumn{2}{c|}{FS L\&F Obstacle} & \multicolumn{3}{c}{Panoptic L\&F [19]} \\
Method & AP $\uparrow$ & FPR95 $\downarrow$ & PQ $\uparrow$ & SQ $\uparrow$& RQ $\uparrow$\\
     \midrule
$ \sigma$-unc.                          & 22.78 & 27.04  & 6.17 & \textbf{78.06} & 7.91\\
naive Beta                  & 12.81 & 23.04 & 0.94 & 62.94 & 1.50 \\
M2F                         & 48.60 & 62.20 & 9.02 & 75.34 & 11.98 \\
\midrule
P2F [ours] & \textbf{66.58} & \textbf{16.84} & \textbf{11.22} & 74.47 & \textbf{15.06} \\
\bottomrule
\end{tabular}
\vskip -0.2cm
\caption{Uncertainty comparison of \ours\ with other naive uncertainties. }
\label{tab:naive_uncertainties}
\vskip -0.2cm
\end{table}  

In \cref{tab:combined_uncertainty}, we evaluate the influence of the individual components in \cref{eq:finalunc}. We evaluate our mask part $p_M^*(h, w)$ and classification part $p_C^*(h, w)$ compared with our \ours{} uncertainty. It can be seen that the class part performs for AP on large shifts from Cityscapes like SMIYC Road Anomaly, but suffers from a high FPR and panoptic detection of L\&F. \ours{} outperforms both for Panoptic L\&F and shows the lowest FPR for Road Anomaly.

\begin{table}[h]
\vskip -0.2cm
\centering
\setlength{\tabcolsep}{4.8pt}
\begin{tabular}{l|cc|ccc}
\toprule
 & \multicolumn{2}{c|}{Road Anomaly} & \multicolumn{3}{c}{Panoptic L\&F \cite{Gasperini2023}} \\
Method & AP $\uparrow$ & FPR95 $\downarrow$ & PQ $\uparrow$ & SQ $\uparrow$& RQ $\uparrow$\\
\midrule
mask part    &  52.02          & 62.96         & 4.89        & 76.43       & 6.40    \\
class part & \textbf{62.94}  &   100.00 &  0.00           & 0.00    & 0.00        \\
\midrule
P2F [ours] & 58.60 & \textbf{46.32} & \textbf{11.22} & 74.47 & \textbf{15.06} \\
\bottomrule
\end{tabular}

\vskip -0.2cm
\caption{Uncertainty comparison of \ours\ using the classical prediction uncertainty and \ours{} uncertainty definition.}
\label{tab:combined_uncertainty}
\vskip -0.2cm
\end{table}

\begin{figure*}[!h]
    \vskip -0.4cm
    \centering
    \includegraphics[width=\linewidth]{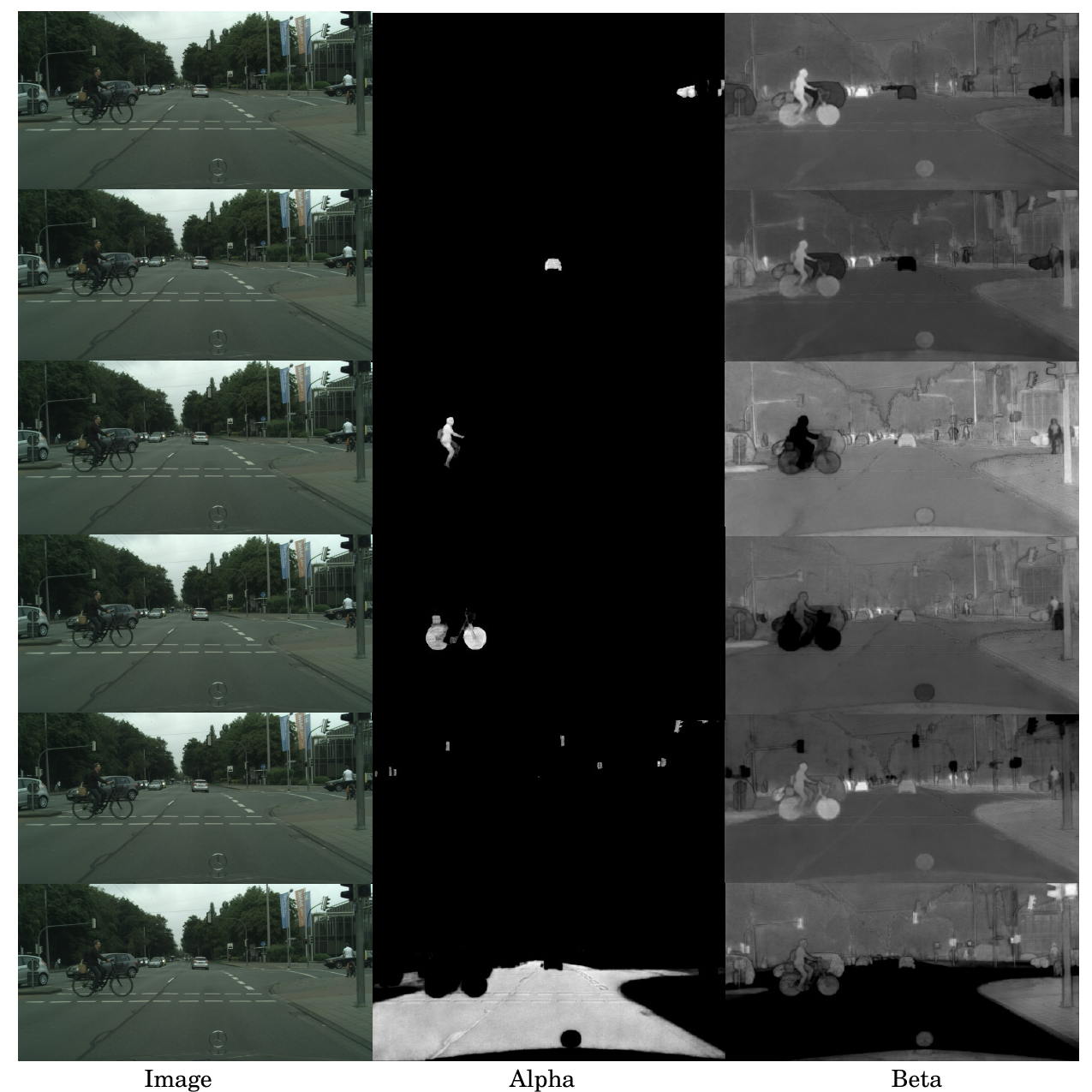}
    \label{fig:prior_vis}
    \caption{Mask Visualization of Alpha and Beta Masks on a single image from Cityscapes \cite{Cordts2016}. Alpha masks represent the positive correspondence of a pixel to a mask, while Beta masks emphasize the negative correspondence.}
\end{figure*}

\section{Additional Qualitative Assessment:}
\label{app:quali}

\subsection{Lost and Found}
In \cref{fig:LandF_open_world_panoptic} we show visual results of open-world panoptic segmentation on the L\&F test set. Anomaly predictions are marked in brown. For all predictions, including the anomaly prediction, different shades mark distinct instances. The L\&F dataset includes the obstacles on the road as anomalies, but also other objects that have not been trained on during training on Cityscapes. These include trash cans, pipes, the back side of traffic signs, and pallets. For anomaly detection, U3HS uses a double threshold strategy where both the classification uncertainty and distances in the embedding are thresholded. For anomaly detection, both predictions need to surpass the individual thresholds. This can lead to not detecting anomalies, as in the second and third row of \cref{fig:LandF_open_world_panoptic}, but also results in less noise in the prediction. Further, the L\&F features uncommon textures of the street, which results in confusion between sidewalk and road or incorrectly detected anomalies, as visible in row one for U3HS and in rows two and four for P2F. 
\subsection{Berkeley Deep Drive}

We also provide further visual results of \ours{} and U3HS~\cite{Gasperini2023} on the BDD~\cite{Yu2020} dataset in \cref{fig:VisualBDD} and on the COCO~\cite{Lin2014} Dataset in \cref{fig:VisualCOCO}.
Compared to Cityscapes, the BDD dataset comprises a higher variety of classes and different scenarios, making the anomaly instance segmentation task more challenging. 
This is demonstrated by the increased false positive rate of both models. 
Nonetheless, they are able to detect anomalies reasonably well in this challenging environment. 
However, the much smaller embedding dimension of U3HS has difficulties in identifying unseen instances, which results in cluttered predictions. This can be seen, for example, in rows one and three in \cref{fig:VisualBDD}. While the explicit regularization of the embeddings prevents uncertainty between instances, \ours{} might predict uncertainties between the transition or regions. Improve the uncertainty instance clustering to reject these remains for future work.  
Additionally, given the high-class imbalance of the dataset, U3HS tends to confuse rare classes with anomalies, for example, the traffic sign pole in row three.
In contrast, \ours\ clearly detects the anomaly as its embeddings are high-dimensional and because of its additional mask supervision signal.
Additionally, U3HS also marks the rider of the bicycles and motorcycles as an anomaly, whereas \ours\ separates well-visible humans, which is a natural phenomenon since human beings are present in the training set.
\vfill\eject

\subsection{COCO}

In comparison to BDD \cite{Yu2020}, the COCO \cite{Lin2014} dataset features less cluttered images but includes a greater variety of classes. The total number of objects per image is lower, and the objects themselves are generally larger.
In this setup, U3HS \cite{Yu2020} and \ours{} perform well on simple images with only one anomaly as well as on more difficult images with several objects and anomaly instances. 
The distance thresholding of U3HS \cite{Gasperini2023} and uncertainty thresholding suppresses uncertainty estimates at the edge of two classes. The pure uncertainty thresholding of \ours{} marks them as an anomaly, as visible in row 2 in \cref{fig:VisualCOCO}. \ours\ consistently shows a superior object segmentation as well as instance separation compared to U3HS \cite{Gasperini2023}. This underlines the effectiveness of our proposed Beta prior and the high quality of the embedding space learned by \ours{}.

\subsection{Mask Visualization}
\label{ap:maskVisu}
In \cref{fig:prior_vis}, we visualize different prior masks $\pmb{\alpha}$ and $\pmb{\beta}$ on a Cityscapes image. The first two rows show detections of individual cars. Notably, the $\pmb{\beta}$-mask tends to suppress other cars less strongly than it does objects from different classes. This underlines the effective instance recognition of P2F for common classes. Rows three and four depict predictions for a human on a bicycle and the bicycle itself. The last two rows correspond to ``stuff" classes; these often cover multiple instances within the same category, such as the traffic lights in row five. Finally, the last row shows the prediction for the street.

\clearpage

{
    \small
    \bibliographystyle{ieeenat_fullname}
    \bibliography{main}
}

\end{document}